\documentclass[letterpaper]{article}

\makeatletter
\let\arxiv@article@maketitle\@maketitle
\let\arxiv@article@makecaption\@makecaption
\let\arxiv@article@section\section
\let\arxiv@article@subsection\subsection
\let\arxiv@article@normalsize\normalsize
\let\arxiv@article@small\small
\let\arxiv@article@footnotesize\footnotesize
\let\arxiv@article@large\large
\let\arxiv@article@Large\Large
\let\arxiv@article@LARGE\LARGE
\makeatother

\usepackage[preprint]{aaai2027}
\usepackage[hyphens]{url}
\usepackage{graphicx}
\usepackage{float}
\usepackage{natbib}
\usepackage{caption}
\usepackage{amsmath}
\usepackage{amssymb}
\usepackage{algorithm}
\usepackage{algorithmic}
\usepackage{booktabs}
\usepackage{multirow}
\usepackage{xcolor}
\definecolor{adalotaccent}{HTML}{1E6075}
\definecolor{adalotcomment}{HTML}{66717A}

\newcommand{\algphase}[1]{%
  \item[]\vspace{2pt}%
  \textcolor{adalotaccent}{\rule{1.5pt}{7pt}\hspace{4pt}\bfseries #1}%
  \vspace{1pt}}

\definecolor{lotaccent}{HTML}{2563EB}
\newcommand{\lotgain}
[1]
{\,{\scriptsize\textcolor{lotaccent}{(#1)}}}
\newcommand{\oursgain}
[1]
{\,{\scriptsize\textcolor{adalotaccent}{(#1)}}}
\newcommand{\method}{AREA}
\newcommand{\methodfull}{Adaptive Relevance-guided Evidence Allocation}
\newcommand{\ind}{\mathbf{1}}

\title{Layers, Sinks, and Scaling: Adaptive Evidence Selection for Multimodal Large Language Models}
\author{\fontsize{12}{14}\selectfont\mbox{Zhenbin Wang, Lei Zhang$^{*}$, Lituan Wang, Wei Huang, Yan Wang, Zhenwei Zhang}}
\affiliations{Sichuan University\\\texttt{wangzhenbin@stu.scu.edu.cn}}

\begin{document}

\maketitle
\begingroup
\renewcommand{\thefootnote}{\fnsymbol{footnote}}
\footnotetext[1]{Lei Zhang is the corresponding author\\
\hspace*{1.8em}Code: \url{https://github.com/wongzbb/AREA}}
\endgroup

\begin{abstract}
Multimodal large language models (MLLMs) can answer knowledge-intensive visual questions by combining visual evidence from images with facts retrieved from external sources. However, MLLMs may overlook relevant evidence in both modalities, attending weakly to the textual sentences or visual regions needed for the correct answer. Recent efforts address this by highlighting retrieved text and marking visual regions before generation, but apply a fixed, one-shot policy that cannot adapt to three sources of variation: whether highlighting is necessary, how much evidence different examples require, and when different textual evidence becomes relevant as the answer unfolds. We introduce \methodfull{} (\method{}), a training-free inference-time method that formulates evidence highlighting as adaptive allocation. \method{} generates a single probe token to read visual and textual relevance from fixed backbone layers, then makes three decisions: \textit{i)} whether to intervene (controlled by natural attention coverage and visual sink contamination), \textit{ii)} how much evidence to expose (determined by relevance entropy), and \textit{iii)} when to refresh text during generation (triggered by causal context-attention peaks). Across four KB-VQA and seven standard multimodal benchmarks with nine frozen MLLM checkpoints, establishes the best performance among training-free highlighting methods.
\end{abstract}

\section{Introduction}

Knowledge-based visual question answering (KB-VQA) requires multimodal large language models (MLLMs) to combine localized visual cues with facts retrieved from external knowledge sources \citep{chen2023infoseek,mensink2023encyclopedic}. Supplying both modalities, however, does not solve the task: the model must still determine which textual statements and image regions are jointly needed to answer the question. Retrieved passages mix critical facts with irrelevant or weakly related content \citep{caffagni2024wikillava}, while images contain many salient objects and regions beyond the answer-bearing area \citep{kang2025seewhat}. Consequently, the model can overlook a required sentence, focus on the wrong visual region, or fail to connect complementary evidence across modalities, producing an incorrect answer even when all required evidence is available.

The challenge therefore lies not only in obtaining relevant evidence, but also in ensuring that the generator uses evidence already present in its input. While retrieval and filtering methods improve which external content reaches the model \citep{hong2025knowledge,yan2024echosight,yang2025omgm,ye2026qkvqa}, a complementary line of work targets evidence utilization at inference time. Recent training-free methods improve evidence utilization through explicit highlighting: SelfElicit highlights relevant context sentences, while Look Twice (LoT) extends the intervention to both retrieved text and visual regions \citep{liu2025selfelicit,morini2026looktwice}. By converting selected evidence into sentence markers and visual crops, such methods make latent evidence explicit without updating model parameters. However, they instantiate highlighting as a fixed, one-shot policy: the same number of sentences and the same visual extent are selected once before answer generation.

\begin{figure}[t]
  \centering
  \includegraphics[width=\columnwidth]{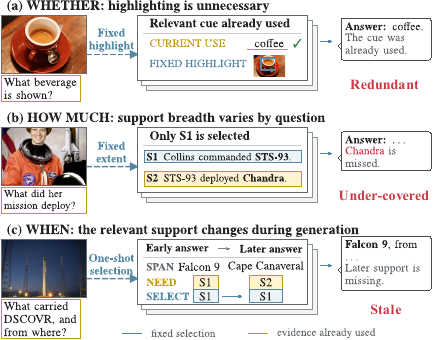}
  \caption{Schematic illustration of three mismatches between fixed, one-shot evidence highlighting and evidence demand. (a) Highlighting is redundant when the model already uses the relevant evidence. (b) A fixed sentence budget can omit part of a multi-sentence evidence chain. (c) Evidence selected before decoding can become stale when later answer tokens require different textual evidence. Blue marks the fixed selection, gold marks evidence already used or required at the corresponding stage, and red labels the resulting mismatch.}
  \label{fig:motivation}
\end{figure}

This design overlooks variation across examples and generation steps, as illustrated in Figure~\ref{fig:motivation}. In some cases, the model already uses the relevant evidence, so additional highlighting is unnecessary. The required evidence also varies in scope: one sentence or a compact image region may suffice for some questions, whereas others depend on multiple sentences or broader visual context. As the answer unfolds, the relevant textual evidence can change, with later tokens depending on context that was not selected before decoding. Static evidence allocation therefore cannot match heterogeneous and time-varying demand. An effective policy must decide whether to intervene, how much evidence to expose, and when to refresh textual evidence.

We introduce \methodfull{} (\method{}), a training-free inference-time method that turns evidence highlighting into adaptive allocation. A single probe token reads visual and textual relevance from fixed, backbone-specific layer groups. \method{} uses the entropy of the textual and visual relevance distributions to set the number of selected sentences and the crop scale. It then gates the two modalities independently: the text gate measures how much attention the selected sentences already receive, while the visual gate measures the fraction of the unfiltered visual relevance assigned to detected sink tokens. During generation, \method{} monitors attention over the original retrieved-context tokens. When its entropy exceeds a threshold computed from earlier decoding steps, the method reselects relevant sentences and appends them at the current decoding position, subject to a fixed refresh budget. Together, these operations adapt whether to intervene, how much evidence to expose, and when to refresh textual evidence, while keeping all model parameters frozen.

%
The main contributions of this work are summarized as follows:
\begin{itemize}
    \item We formalize adaptive evidence allocation for frozen MLLMs through three decisions: whether to mark selected context sentences and whether to add a visual crop, how many sentences to mark and what spatial extent the crop should cover, and at which decoding steps to reselect and append marked sentences.
    \item We develop \method{}, a training-free method for frozen MLLMs that obtains layer-resolved visual and textual relevance in one probe pass, uses it to set sentence count and crop extent, gates text and vision with selected-sentence attention surprisal and raw visual sink mass, and triggers text refresh from decoding-time context attention under a fixed budget.
    \item We evaluate \method{} with nine frozen MLLM checkpoints on four KB-VQA and seven standard multimodal benchmarks, where it consistently establishes state-of-the-art performance across both knowledge-intensive and standard multimodal settings under a frozen-backbone inference protocol.
\end{itemize}

\section{Related Work}

\subsection{KB-VQA and Multimodal Retrieval}

KB-VQA combines image understanding with facts from external sources. Benchmarks like Encyclopedic VQA, InfoSeek, and ViQuAE test whether models can answer questions about fine-grained entities, unseen knowledge, or facts requiring multi-step reasoning \citep{mensink2023encyclopedic,chen2023infoseek,lerner2022viquae}. Systems that perform well on these tasks typically improve the upstream evidence pipeline: hierarchical retrieval over structured knowledge \citep{caffagni2024wikillava}, multimodal reranking to prioritize relevant passages \citep{yan2024echosight,yang2025omgm}, learned retrieval-relevance decisions \citep{cocchi2025reflectiva}, reasoning-augmented retrieval \citep{compagnoni2026reag}, question-focused filtering \citep{ye2026qkvqa}, and multimodal knowledge graph integration \citep{yuan2026mkgrag}.
These methods determine which external content reaches the model. \method{} addresses the complementary question: given that relevant evidence is available in the input, how should it be presented to ensure the frozen generator uses it? This distinction permits controlled comparisons under identical retrieval, isolating evidence utilization from retrieval quality.

\subsection{Inference-Time Evidence Highlighting}

Training-free evidence highlighting provides a direct approach to improving evidence utilization without changing the retrieval pipeline or updating model parameters. SelfElicit uses model-derived relevance to select and mark context sentences, while LoT extends this approach to MLLMs by combining textual highlighting with visual localization \citep{liu2025selfelicit,morini2026looktwice}. Together, these methods show that relevance signals from a frozen model can guide how available evidence is presented during inference. Existing approaches, however, make the intervention in a single pre-decoding step under a predetermined allocation policy. \method{} advances this direction by independently adapting the intervention for text and vision, scaling the exposed evidence to each example, and refreshing textual evidence as generation unfolds.

\section{Method}

\subsection{Problem Setup and Overview}

Given an image $I$, a question $Q$, and a candidate context $C$ of $N_C$ tokens, \method{} generates an answer with a frozen MLLM $\mathcal{F}_\theta$; vision-only tasks have $N_C=0$. The visual front end maps $I$ to $N_V=G^2$ spatial patch tokens arranged on a square $G\times G$ grid. The decoder has $L$ layers, $N_H$ attention heads per layer, and hidden width $d$.
From the unhighlighted tokenized prompt, $\mathcal{F}_\theta$ generates exactly one probe token. Let $S_p$ denote the resulting sequence length and $i_{\rm p}$ the final position occupied by that token. For layer $\ell\in\{1,\ldots,L\}$ and head $h\in\{1,\ldots,N_H\}$, the probe pass exposes post-softmax causal attention $\mathbf{A}_p^{\ell,h}\in[0,1]^{S_p\times S_p}$ and hidden states $\mathbf{H}_p^\ell\in\mathbb{R}^{S_p\times d}$. An entry $\mathbf{A}_p^{\ell,h}[r,s]$ gives attention from query position $r$ to key position $s$. Tokenization maps visual patch $v$ to its absolute prompt position $\chi_{\rm vis}(v)$ and, when $N_C>0$, original context token $j$ to $\chi_{\rm ctx}(j)$.

Our objective is to allocate the available visual and textual evidence according to the evidence demand of each example and generation step. \method{} therefore decides \textit{whether} to intervene in each modality, \textit{how much} evidence to expose through the number of selected sentences and the visual crop extent, and \textit{when} to refresh textual evidence during decoding, while keeping $\mathcal{F}_\theta$ frozen.

\begin{figure*}[t]
  \centering
  \includegraphics[width=0.95\textwidth]{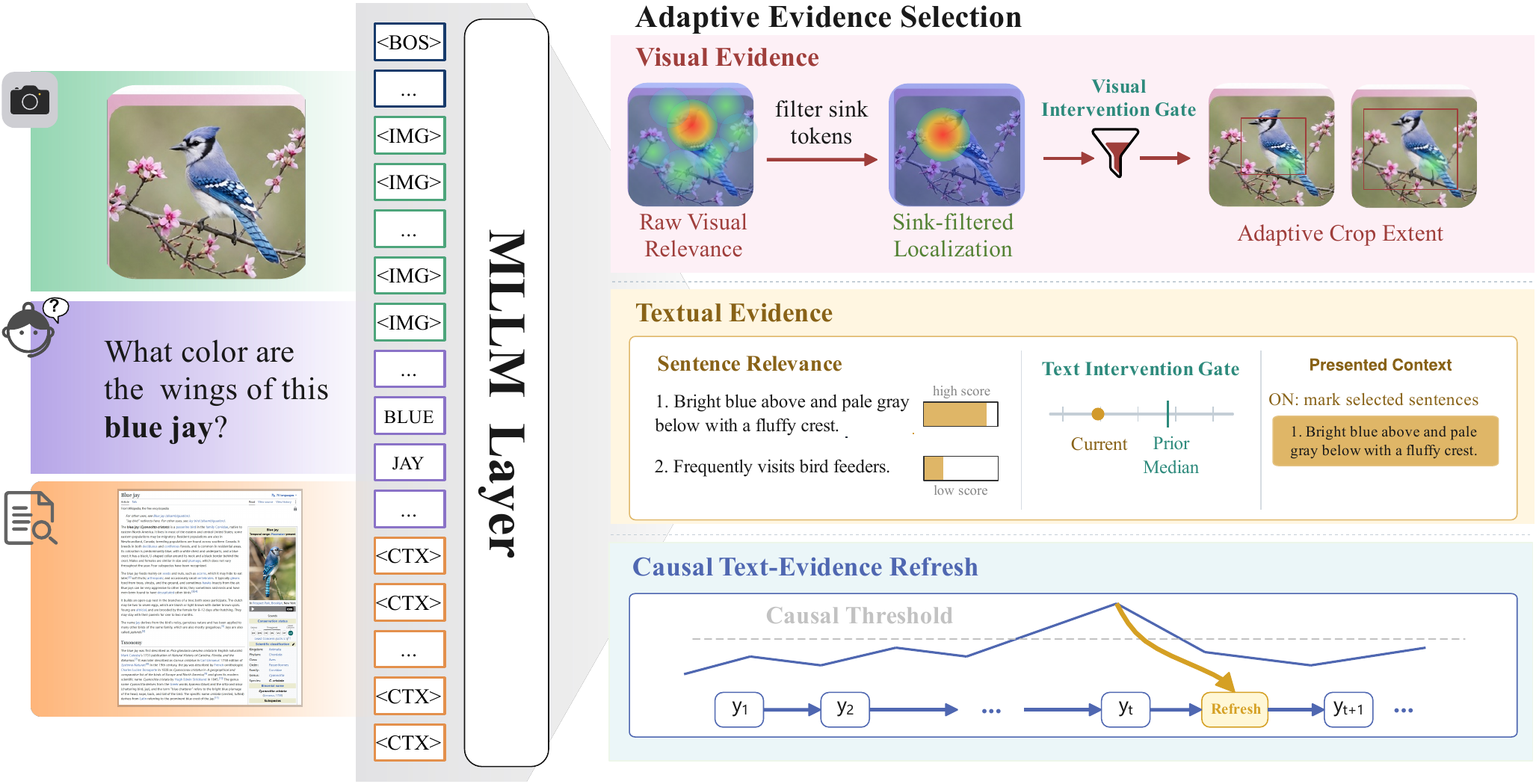}
  \caption{Overview of \method{}. One probe token reads visual and textual relevance from fixed, backbone-specific layer groups of a frozen MLLM. Independent gates use selected-sentence attention coverage and raw visual sink mass to decide whether to intervene, while the sentence-level and sink-filtered spatial relevance distributions set the sentence count and crop extent. After warmup, both gates use prior-sample medians and update their histories only after deciding. During the same autoregressive stream, strictly causal peaks in context-attention entropy trigger bounded textual refresh; visual evidence and the initial gates remain fixed.}
  \label{fig:framework}
\end{figure*}

Figure~\ref{fig:framework} connects Layer-Resolved Evidence Readout to Entropy-Calibrated Evidence Scaling and Modality-Specific Intervention Gating, followed by Causal Text-Evidence Refresh during generation. Algorithm~\ref{alg:area} presents the complete inference sequence.

\subsection{Layer-Resolved Evidence Readout}

We use nonempty, fixed, backbone-specific layer groups $\mathcal{L}_{\rm vis},\mathcal{L}_{\rm txt}\subseteq\{1,\ldots,L\}$ for visual and textual relevance readout, together with nonempty fixed sink dimensions $\mathcal{D}_{\rm sink}\subseteq\{1,\ldots,d\}$ for activation-based visual sink detection \citep{kang2025seewhat}. The fixed question parser identifies the target-object phrase; let $\mathcal{O}$ denote the nonempty set of its absolute prompt positions. The same probe pass produces raw visual relevance for each patch $v$ and, when context is present, textual relevance for each original context token $j$:
\begin{equation}
\begin{aligned}
a_{{\rm vis},v}^{\rm raw}
&=\frac{\displaystyle
\sum_{o\in\mathcal{O}}\sum_{\ell\in\mathcal{L}_{\rm vis}}
\sum_{h=1}^{N_H}\mathbf{A}_p^{\ell,h}[o,\chi_{\rm vis}(v)]}
{|\mathcal{O}|\,|\mathcal{L}_{\rm vis}|N_H},\\
a_{{\rm txt},j}
&=\frac{\displaystyle
\sum_{\ell\in\mathcal{L}_{\rm txt}}\sum_{h=1}^{N_H}
\mathbf{A}_p^{\ell,h}[i_{\rm p},\chi_{\rm ctx}(j)]}
{|\mathcal{L}_{\rm txt}|N_H}.
\end{aligned}
\label{eq:relevance}
\end{equation}
Collecting these scores yields the raw visual relevance vector $\mathbf{a}_{\rm vis}^{\rm raw}\in\mathbb{R}_{\geq0}^{N_V}$ and textual relevance vector $\mathbf{a}_{\rm txt}\in\mathbb{R}_{\geq0}^{N_C}$. Thus the visual branch reads object-to-patch attention, while the textual branch reads probe-to-context attention. Both directly average the original post-softmax weights over their fixed layers and all heads.

\paragraph{Visual sink filtering.}
Visual sink tokens concentrate activation in a small set of hidden dimensions and can distort the relevance map used to localize the target object. We score this concentration by normalizing the strongest activation in $\mathcal{D}_{\rm sink}$ by the root mean square across all $d$ hidden dimensions:
\begin{equation}
s_{{\rm sink},v}
=\frac{1}{|\mathcal{L}_{\rm vis}|}
\sum_{\ell\in\mathcal{L}_{\rm vis}}
\frac{\displaystyle\max_{r\in\mathcal{D}_{\rm sink}}
|\mathbf{H}_p^\ell[\chi_{\rm vis}(v),r]|}
{\displaystyle\sqrt{\frac{1}{d}\sum_{q=1}^{d}
\bigl(\mathbf{H}_p^\ell[\chi_{\rm vis}(v),q]\bigr)^2}}.
\label{eq:sink}
\end{equation}
For visual sink threshold $\tau$, tokens satisfying $s_{{\rm sink},v}>\tau$ form $\mathcal{I}_{\rm sink}$. We define $\mathbf{a}_{\rm vis}=(a_{{\rm vis},1},\ldots,a_{{\rm vis},N_V})\in\mathbb{R}_{\geq0}^{N_V}$ by setting $a_{{\rm vis},v}=0$ inside $\mathcal{I}_{\rm sink}$ and $a_{{\rm vis},v}=a_{{\rm vis},v}^{\rm raw}$ otherwise. This filtered vector determines crop location and scale, and the corresponding raw sink mass controls the visual intervention gate.

\paragraph{Spatial localization.}
We reshape $\mathbf{a}_{\rm vis}$ in the visual processor's patch order and divide each cell by the sum over all $G^2$ cells, obtaining the spatial distribution $\widetilde{M}_{u,w}$ for $u,w\in\{0,\ldots,G-1\}$. Its center and spread are
\begin{equation}
\begin{aligned}
\mu_x&{=}\sum_{u,w}w\widetilde{M}_{u,w}, \quad
\sigma_x{=}\sqrt{(\sum_{u,w}(w-\mu_x)^2
\widetilde{M}_{u,w})},\\
\mu_y&{=}\sum_{u,w}u\widetilde{M}_{u,w}, \quad
\sigma_y{=}\sqrt{(\sum_{u,w}(u-\mu_y)^2
\widetilde{M}_{u,w})}.
\end{aligned}
\label{eq:spatial}
\end{equation}
These moments locate the crop and define its horizontal and vertical extent before adaptive scaling.

\subsection{Entropy-Calibrated Evidence Scaling}

\paragraph{Text scaling.}
For $N_C>0$, a sentence segmenter maps the original context to token sets $\{\mathcal{S}_m\}_{m=1}^{N_S}$. We form the sentence distribution $\mathbf{p}=(p_1,\ldots,p_{N_S})$ by setting $p_m\propto |\mathcal{S}_m|^{-1}\sum_{j\in\mathcal{S}_m}a_{{\rm txt},j}$ and normalizing it so that $\sum_{m=1}^{N_S}p_m=1$. Let $H(\mathbf{p})=-\sum_{m=1}^{N_S}p_m\log_2p_m$. We select the sentence indices
\begin{equation}
\mathcal{J}_{\rm txt}
=\operatorname{TopK}\!\left(
\mathbf{p},
\min\!\left(\left\lceil2^{H(\mathbf{p})}\right\rceil,k_{\max}\right)
\right).
\label{eq:textscale}
\end{equation}
Eq.~\eqref{eq:textscale} converts relevance dispersion into an evidence budget: low entropy indicates concentrated evidence and selects fewer sentences, whereas high entropy indicates distributed evidence and selects more. The effective count $2^{H(\mathbf{p})}$, capped at $k_{\max}$, sets this budget before $\operatorname{TopK}$ retains the most relevant sentences.

\paragraph{Visual scaling.}
The spatial entropy controls crop extent:
\begin{equation}
\begin{aligned}
&\mathcal{H}_{\rm vis}
=-\sum_{u,w}\widetilde{M}_{u,w}\log_2\widetilde{M}_{u,w},\\
&\beta
=\operatorname{clip}\!\left(
\frac{1}{2}\sqrt{\frac{2^{\mathcal{H}_{\rm vis}}}{\sigma_x\sigma_y}},
\beta_{\min},\beta_{\max}\right),\\
&\mathbf{b}_{\rm pix}
=\Pi_{\rm pix}\!\left(
\mu_x{-}\beta\sigma_x,\mu_y{-}\beta\sigma_y,
\mu_x{+}\beta\sigma_x,\mu_y{+}\beta\sigma_y\right).
\end{aligned}
\label{eq:visscale}
\end{equation}
The fixed map $\Pi_{\rm pix}$ converts grid coordinates to pixels and clips the box to the image boundary. Because $2^{\mathcal{H}_{\rm vis}}$ is the effective number of grid cells carrying relevance, the unconstrained scale satisfies $(2\beta\sigma_x)(2\beta\sigma_y)=2^{\mathcal{H}_{\rm vis}}$, tying crop area to spatial relevance dispersion. The centroid and axis-wise spreads retain the map's location and shape, so concentrated maps yield tight crops while diffuse maps yield broader ones.

\subsection{Modality-Specific Intervention Gating}
\label{sec:gating}

The preceding scaling stage determines the candidate text and crop, but presenting them unconditionally would ignore whether intervention is needed. Because the model may already use the relevant evidence and intervention demand can differ by modality, \method{} gates text and vision independently. The text score measures how little natural attention reaches the selected sentences, while the visual score is the fraction of raw relevance assigned to sink-flagged patches:
\begin{equation}
\begin{aligned}
u_{\rm txt}
&=-\log_2\!\left(
\frac{\sum_{m\in\mathcal{J}_{\rm txt}}
\sum_{j\in\mathcal{S}_m}a_{{\rm txt},j}}
{\sum_{j=1}^{N_C}a_{{\rm txt},j}}
+\varepsilon\right),\\
u_{\rm vis}
&=\frac{\sum_{v\in\mathcal{I}_{\rm sink}}
a_{{\rm vis},v}^{\rm raw}}
{\sum_{v=1}^{N_V}a_{{\rm vis},v}^{\rm raw}}.
\end{aligned}
\label{eq:gatescores}
\end{equation}
A high $u_{\rm txt}$ indicates that the selected sentences are underused, and a high $u_{\rm vis}$ indicates stronger sink contamination in the raw localization signal.

For sample $n$ and modality $x\in\{{\rm txt},{\rm vis}\}$, let $m_x^{(<n)}$ be the median of duplicate-retaining scores from earlier eligible samples in the current dataset--backbone stream. Vision is always eligible, whereas text is eligible only when $N_C>0$; thus $z_{\rm txt}^{(n)}=0$ when $N_C=0$. For each eligible modality, we set $z_x^{(n)}=1$ during the first $W_{\rm warm}$ samples and $z_x^{(n)}=\ind[u_x\ge m_x^{(<n)}]$ thereafter.
The two decisions are independent and use only preceding samples. After both decisions, we append the current eligible scores to their histories; histories reset between datasets and backbones.
When $z_{\rm txt}=1$, $C^\star$ is $C$ with the sentences in $\mathcal{J}_{\rm txt}$ wrapped by text-evidence markers, otherwise $C^\star=C$. When $z_{\rm vis}=1$, the image payload $\mathcal{I}^\star$ is the marker-wrapped crop defined by $\mathbf{b}_{\rm pix}$ alone, otherwise $\mathcal{I}^\star$ contains the original image.

\subsection{Causal Text-Evidence Refresh}

As generation progresses, evidence needs can shift beyond the initial selection. \method{} therefore monitors context-attention dispersion and reselects and appends textual evidence when the current value exceeds a threshold determined only by earlier decoding steps, subject to a fixed refresh budget.

The initial generation prompt $\mathcal{P}_1$ is tokenized over $(\mathcal{I}^\star,Q,C^\star)$. When $N_C>0$, tokenization supplies the order-preserving map $j\mapsto\iota(j)$ from each original context token to its prompt position; evidence-marker tokens are excluded. $R_{\max}$ caps refreshes, and $r_t$ counts those completed through step $t$, with $r_0=0$.

For $t\geq1$, $\mathcal{P}_t$ is the cached prefix before decoding $y_t$. Once committed and cached, $y_t$ occupies $i_t=|\mathcal{P}_t|+1$; let $\mathbf{A}_t^{\ell,h}[i_t,:]\in[0,1]^{i_t}$ denote its post-softmax causal-attention row. Applying the textual readout of Eq.~\eqref{eq:relevance} to this row at keys $\iota(j)$ yields $\mathbf{a}_{\rm txt}^{(t)}\in\mathbb{R}_{\geq0}^{N_C}$. We normalize these scores and compute their entropy:
\begin{equation}
\pi_j^{(t)}=\frac{a_{{\rm txt},j}^{(t)}}{\sum_{q=1}^{N_C}a_{{\rm txt},q}^{(t)}},\quad
e_t=-\sum\nolimits_{j=1}^{N_C}\pi_j^{(t)}\log_2\pi_j^{(t)}.
\label{eq:stepentropy}
\end{equation}
The weights $\pi_j^{(t)}$ cover only original context tokens; evidence markers and appended reminders are excluded. Their entropy $e_t$ is low for concentrated attention and high for dispersed attention, serving as a step-wise evidence-demand proxy.

For $t>1$, $\bar e_{t-1}$ and $\sigma_{e,t-1}$ are the mean and population standard deviation of $(e_1,\ldots,e_{t-1})$; $\kappa\geq0$ controls peak sensitivity. The binary trigger is
\begin{equation}
\gamma_t{=}
\begin{cases}
\ind[e_t>\bar e_{t{-}1}{+}\kappa\sigma_{e,t{-}1}],
&t{>}1\ \land\ r_{t{-}1}{<}R_{\max},\\
0,&\text{otherwise}.
\end{cases}
\label{eq:trigger}
\end{equation}
The strict test excludes $e_t$ from its own threshold. The first eligible trigger is therefore $t=2$, where the one-value population standard deviation is zero.

When $\gamma_t=1$, substituting $\mathbf{a}_{\rm txt}^{(t)}$ for $\mathbf{a}_{\rm txt}$ in the sentence aggregation and Eq.~\eqref{eq:textscale} yields $\mathcal{J}_{\rm txt}^{(t)}$. We serialize these sentences in context order and define $R_t$ as the resulting token block, enclosed once by the text-evidence markers. With brackets denoting token-sequence concatenation, the cached prefix and refresh count evolve as
\begin{equation}
\begin{aligned}
\mathcal{P}_{t+1}
&=\begin{cases}
[\mathcal{P}_t;y_t;R_t],&\gamma_t=1,\\
[\mathcal{P}_t;y_t],&\gamma_t=0,
\end{cases}&
r_t&=r_{t-1}+\gamma_t.
\end{aligned}
\label{eq:refresh}
\end{equation}
Appending $R_t$ at the decoding frontier preserves $\iota$ and existing key--value cache entries; $e_t$ then enters the history used at step $t+1$. The mechanism refreshes only textual evidence and remains active when $z_{\rm txt}=0$.

\FloatBarrier
\subsection{Inference Protocol}

\begin{algorithm}[H]
\caption{\method{} Inference}
\label{alg:area}
\small
\begin{algorithmic}[1]
\REQUIRE $I,Q,C$; frozen $\mathcal{F}_\theta$; state $\mathcal{U}_{\rm txt},\mathcal{U}_{\rm vis},n_{\rm seen}$
\algphase{Layer-Resolved Evidence Readout}
\STATE Generate exactly one probe token; expose $\mathbf{A}_p^{\ell,h},\mathbf{H}_p^\ell$
\STATE Resolve $\mathcal{O},\chi_{\rm vis}$ and, when $N_C>0$, $\chi_{\rm ctx}$
\STATE Compute $\mathbf{a}_{\rm vis}^{\rm raw}$ and, when $N_C>0$, $\mathbf{a}_{\rm txt}$
\STATE Compute sink scores and $\mathcal{I}_{\rm sink}=\{v:s_{{\rm sink},v}>\tau\}$
\STATE Form $\mathbf{a}_{\rm vis}$, $\widetilde{M}$, and its spatial moments
\algphase{Entropy-Calibrated Evidence Scaling}
\IF{$N_C>0$}
  \STATE Form $\mathbf{p}$ and select $\mathcal{J}_{\rm txt}$ by Eq.~\eqref{eq:textscale}
\ENDIF
\STATE Compute $\mathcal{H}_{\rm vis},\beta,\mathbf{b}_{\rm pix}$ by Eq.~\eqref{eq:visscale}
\algphase{Modality-Specific Intervention Gating}
\STATE Compute $u_{\rm vis}$ and, if $N_C>0$, $u_{\rm txt}$
\STATE Set eligible $z_x$ from warmup or prior medians \COMMENT{prior only}
\STATE Append each eligible $u_x$; increment $n_{\rm seen}$ \COMMENT{after both gates}
\STATE Form the conditional payloads $\mathcal{I}^\star$ and $C^\star$
\STATE Tokenize $\mathcal{P}_1$ and, if $N_C>0$, obtain $\iota$
\algphase{Causal Text-Evidence Refresh}
\STATE Set $r_0\leftarrow0$ and initialize an empty entropy history
\FOR{$t=1,2,\ldots$ until generation terminates}
  \STATE Decode, commit, and cache $y_t$
  \IF{$N_C>0$}
    \STATE Read $\mathbf{a}_{\rm txt}^{(t)}$ through $\iota$; compute $\pi_j^{(t)}$ and $e_t$
    \STATE Set $\gamma_t$ by the strict prior-history test in Eq.~\eqref{eq:trigger}
    \IF{$\gamma_t=1$}
      \STATE Reselect $\mathcal{J}_{\rm txt}^{(t)}$ and form $R_t$ \COMMENT{text only}
    \ENDIF
    \STATE Update $(\mathcal{P}_{t+1},r_t)$ by Eq.~\eqref{eq:refresh}
    \STATE Append $e_t$ to the entropy history \COMMENT{after decision}
  \ENDIF
\ENDFOR
\RETURN generated answer tokens
\end{algorithmic}
\end{algorithm}

\section{Experiments}

\subsection{Experimental Setup}

\begin{table*}
[t]
\centering
\small
\setlength{\tabcolsep}{4.0pt}
\renewcommand{\arraystretch}{0.88}

\begin{tabular}{cllcccccccc}
\toprule
& Backbone & Method
& \multicolumn{2}{c}{E-VQA}
& \multicolumn{3}{c}{InfoSeek}
& OVEN & ViQuAE & Avg \\
\cmidrule(lr){4-5}
\cmidrule(lr){6-8}
& & & SH & All & U-Q & U-E & All & All & All & \\
\midrule

\multirow{12}{*}{\rotatebox
[origin=c]
{90}{Small}}
& \multirow{3}{*}{Qwen2-VL-2B}
& Base
& 17.5 & 16.0 & 5.3 & 5.6 & 5.4 & 1.2 & 15.0 & 10.2 \\
& & \hspace{0.6em}\bfseries +LoT
& 18.2 & 16.2 & 10.5 & 10.1 & 10.3 & 2.8 & 18.4
& 11.9\lotgain{+1.7} \\
& & \hspace{0.6em}\textcolor{adalotaccent}{\bfseries +\method{} (Ours)}
& 18.7 & 16.7 & 11.4 & 10.0 & 10.9 & 3.4 & 19.2
& 12.6\oursgain{+2.4} \\

\addlinespace
[0.5pt]

& \multirow{3}{*}{Qwen2.5-VL-3B}
& Base
& 30.1 & 27.8 & 22.6 & 22.2 & 22.4 & 11.5 & 22.9 & 21.2 \\
& & \hspace{0.6em}\bfseries +LoT
& 32.9 & 30.4 & 25.4 & 25.1 & 25.2 & 18.3 & 27.8
& 25.5\lotgain{+4.3} \\
& & \hspace{0.6em}\textcolor{adalotaccent}{\bfseries +\method{} (Ours)}
& 33.4 & 31.0 & 25.9 & 24.9 & 25.6 & 19.0 & 28.5
& 26.0\oursgain{+4.8} \\

\addlinespace
[0.5pt]

& \multirow{3}{*}{Qwen3-VL-4B}
& Base
& 35.0 & 32.8 & 27.9 & 28.6 & 28.3 & 25.1 & 34.5 & 30.2 \\
& & \hspace{0.6em}\bfseries +LoT
& 37.3 & 34.8 & 30.8 & 30.2 & 30.5 & 24.5 & 36.3
& 31.5\lotgain{+1.3} \\
& & \hspace{0.6em}\textcolor{adalotaccent}{\bfseries +\method{} (Ours)}
& 37.9 & 35.4 & 31.4 & 30.0 & 30.9 & 25.1 & 37.0
& 32.1\oursgain{+1.9} \\

\addlinespace
[0.5pt]

& \multirow{3}{*}{InternVL3.5-4B}
& Base
& 29.1 & 26.2 & 28.8 & 29.0 & 28.9 & 10.8 & 36.4 & 25.6 \\
& & \hspace{0.6em}\bfseries +LoT
& 31.7 & 28.7 & 33.3 & 33.1 & 33.2 & 11.5 & 45.6
& 29.8\lotgain{+4.2} \\
& & \hspace{0.6em}\textcolor{adalotaccent}{\bfseries +\method{} (Ours)}
& 32.3 & 29.3 & 33.9 & 32.9 & 33.6 & 12.2 & 46.5
& 30.4\oursgain{+4.8} \\

\midrule

\multirow{12}{*}{\rotatebox
[origin=c]
{90}{Medium}}
& \multirow{3}{*}{Qwen2-VL-7B}
& Base
& 25.6 & 22.9 & 24.2 & 24.7 & 24.4 & 11.1 & 33.0 & 22.9 \\
& & \hspace{0.6em}\bfseries +LoT
& 28.7 & 25.6 & 29.8 & 30.1 & 29.9 & 16.6 & 40.8
& 28.2\lotgain{+5.3} \\
& & \hspace{0.6em}\textcolor{adalotaccent}{\bfseries +\method{} (Ours)}
& 29.3 & 26.2 & 30.5 & 29.9 & 30.2 & 17.2 & 41.6
& 28.8\oursgain{+5.9} \\

\addlinespace
[0.5pt]

& \multirow{3}{*}{Qwen2.5-VL-7B}
& Base
& 32.1 & 30.1 & 23.9 & 25.2 & 24.5 & 21.1 & 36.3 & 28.0 \\
& & \hspace{0.6em}\bfseries +LoT
& 33.6 & 31.4 & 24.9 & 25.4 & 25.1 & 21.2 & 38.6
& 29.1\lotgain{+1.1} \\
& & \hspace{0.6em}\textcolor{adalotaccent}{\bfseries +\method{} (Ours)}
& 34.0 & 31.8 & 25.5 & 25.2 & 25.4 & 21.8 & 39.4
& 29.6\oursgain{+1.6} \\

\addlinespace
[0.5pt]

& \multirow{3}{*}{Qwen3-VL-8B}
& Base
& 36.6 & 35.0 & 29.1 & 30.4 & 29.7 & 17.7 & 43.7 & 31.5 \\
& & \hspace{0.6em}\bfseries +LoT
& 38.6 & 36.4 & 33.2 & 32.5 & 32.8 & 19.6 & 51.0
& 35.0\lotgain{+3.5} \\
& & \hspace{0.6em}\textcolor{adalotaccent}{\bfseries +\method{} (Ours)}
& 39.1 & 36.9 & 33.8 & 32.3 & 33.1 & 20.2 & 51.8
& 35.5\oursgain{+4.0} \\

\addlinespace
[0.5pt]

& \multirow{3}{*}{InternVL3.5-8B}
& Base
& 31.4 & 29.0 & 29.5 & 29.9 & 29.7 & 17.7 & 44.2 & 30.2 \\
& & \hspace{0.6em}\bfseries +LoT
& 33.2 & 30.7 & 32.1 & 32.5 & 32.3 & 17.6 & 54.2
& 33.7\lotgain{+3.5} \\
& & \hspace{0.6em}\textcolor{adalotaccent}{\bfseries +\method{} (Ours)}
& 33.8 & 31.3 & 32.8 & 32.3 & 32.6 & 18.3 & 55.0
& 34.3\oursgain{+4.1} \\

\midrule

\multirow{3}{*}{\rotatebox
[origin=c]
{90}{Large}}
& \multirow{3}{*}{Qwen2.5-VL-32B}
& Base
& 35.3 & 33.7 & 26.7 & 26.1 & 26.4 & 13.3 & 37.9 & 27.8 \\
& & \hspace{0.6em}\bfseries +LoT
& 37.7 & 36.0 & 31.2 & 26.2 & 29.6 & 14.2 & 46.0
& 31.5\lotgain{+3.7} \\
& & \hspace{0.6em}\textcolor{adalotaccent}{\bfseries +\method{} (Ours)}
& 38.2 & 36.5 & 31.8 & 26.0 & 30.1 & 14.9 & 46.8
& 32.1\oursgain{+4.3} \\

\bottomrule
\end{tabular}

\caption{Performance on KB-VQA benchmarks. Base denotes the same checkpoint without evidence highlighting. SH, U-Q, and U-E denote Single-Hop, Unseen-Q, and Unseen-E; Avg is computed over the four All columns. Colored values in parentheses indicate absolute improvements over Base.}
\label{tab:kbvqa_main}
\end{table*}

\begin{figure}[t]
  \centering
  \includegraphics[width=1.0\linewidth]{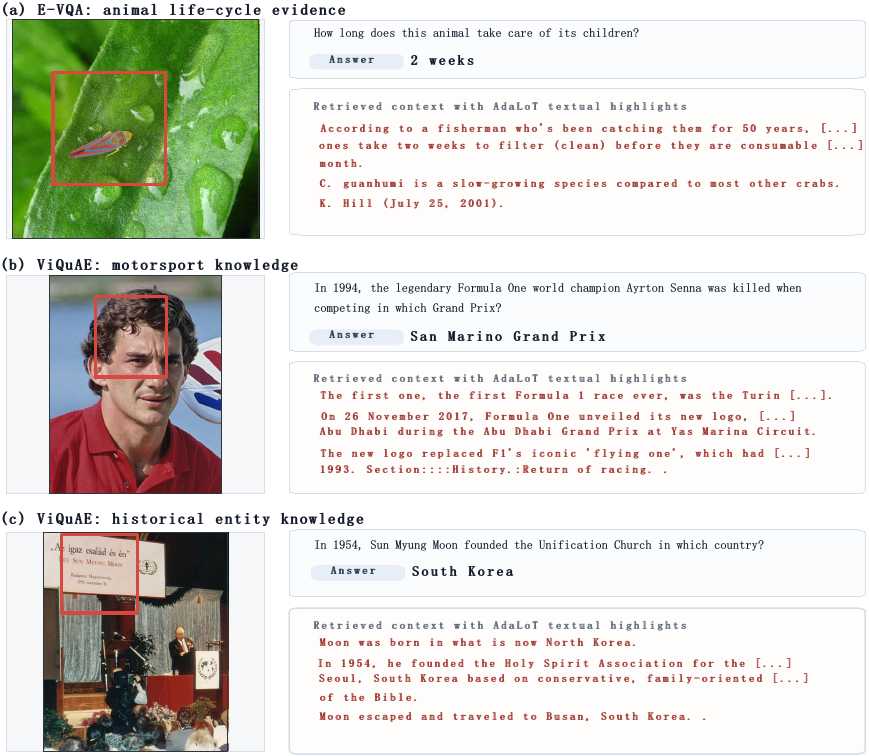}
  \caption{Qualitative examples of AREA on E-VQA and ViQuAE. Each row shows the original image with the AREA-induced visual bounding box and the retrieved textual context with AREA-selected sentences highlighted. The examples illustrate how AREA jointly localizes visual evidence and emphasizes relevant textual evidence for knowledge-intensive visual question answering. Best viewed when zoomed in.}
  \label{fig:area-qualitative}
\end{figure}

\paragraph{Datasets and metrics.}
We follow the evaluation protocol of LoT \citep{morini2026looktwice}. The knowledge-based visual question answering (KB-VQA) suite comprises Encyclopedic VQA (E-VQA) \citep{mensink2023encyclopedic}, InfoSeek \citep{chen2023infoseek}, Open-domain Visual Entity Recognition (OVEN) \citep{hu2023oven}, and ViQuAE \citep{lerner2022viquae}. We report Single-Hop and All accuracy on the E-VQA test set; Unseen-Q, Unseen-E, and All accuracy on the InfoSeek validation set; validation accuracy on OVEN; and exact match on the ViQuAE test set. The standard suite contains RealWorldQA \citep{xai2024grok}, Vstar \citep{wu2024vstar}, TextVQA \citep{singh2019textvqa}, ChartQA \citep{masry2022chartqa}, OCRBench \citep{liu2024ocrbench}, POPE \citep{li2023pope}, and AMBER-D \citep{wang2023amber}. We use the Cambrian-1 evaluation suite \citep{tong2024cambrian} except for AMBER-D, which uses its official discriminative-accuracy pipeline.

\paragraph{Backbones and protocol.}
We evaluate nine frozen MLLM checkpoints: Qwen2-VL-2B and Qwen2-VL-7B \citep{wang2024qwen2vl}; Qwen2.5-VL-3B, Qwen2.5-VL-7B, and Qwen2.5-VL-32B \citep{bai2025qwen}; Qwen3-VL-4B and Qwen3-VL-8B \citep{bai2025qwen3vl}; and InternVL3.5-4B and InternVL3.5-8B \citep{wang2025internvl35}. For KB-VQA, we retrieve three Wikipedia entities with the same EVA-CLIP and FAISS pipeline used by LoT \citep{sun2024evaclip,johnson2019faiss}. Each LoT--\method{} pair uses the same checkpoint, retrieved entities, task instruction, decoding configuration, and evaluator. LoT selects one sentence, fixes the visual scale to $\beta=2$, and always applies its intervention. \method{} uses $k_{\max}=3$, $(\beta_{\min},\beta_{\max})=(1.5,2.5)$, $\kappa=1$, $R_{\max}=4$ and $W_{\rm warm}=50$.

\subsection{Results on KB-VQA}

Table~1 reports the results on four KB-VQA benchmarks. AREA outperforms LoT for all nine checkpoints, improving the four-dataset average by approximately 0.6 points across backbones. This consistency across model families and scales suggests that the gains arise from better matching the intervention to each example, rather than from a particular architecture. Unlike LoT, which always exposes a fixed crop and a single sentence, AREA avoids redundant interventions when the relevant evidence is already well covered and expands the evidence scope when relevance is more dispersed.

The gains are particularly clear on OVEN and ViQuAE, where answering often requires identifying a visual entity and connecting it to a specific retrieved fact. Adaptive localization can suppress unrelated visual regions, while flexible sentence allocation reduces the risk of omitting supporting information. On InfoSeek, the improvements are mainly observed on Unseen-Q and the overall score, whereas Unseen-E changes only marginally and occasionally decreases. We conjecture that unseen-entity questions are more strongly limited by entity recognition and retrieval quality; when the correct entity or passage is absent, improving evidence presentation alone provides limited benefit.

\begin{table*}
[t]
\centering
\small
\setlength{\tabcolsep}{1.6pt}
\renewcommand{\arraystretch}{0.88}

\begin{tabular}{cllccccccc}
\toprule
& Backbone & Method
& \multicolumn{5}{c}{Vision-Centric/OCR and Chart}
& \multicolumn{2}{c}{Hallucination} \\
\cmidrule(lr){4-8}
\cmidrule(lr){9-10}
& & & RealWorldQA & V-Star & TextVQA & ChartQA & OCRBench & POPE & AMBER-D \\
\midrule

\multirow{12}{*}{\rotatebox
[origin=c]
{90}{Small}}
& \multirow{3}{*}{Qwen2-VL-2B}
& Base
& 54.8 & 46.6 & 72.0 & 73.4 & 74.6 & 88.4 & 42.1 \\
& & \hspace{0.6em}\bfseries +LoT
& 56.7\lotgain{+1.9}
& 53.9\lotgain{+7.3}
& 74.9\lotgain{+2.9}
& 73.2\lotgain{-0.2}
& 74.0\lotgain{-0.6}
& 88.6\lotgain{+0.2}
& 75.0\lotgain{+32.9} \\
& & \hspace{0.6em}\textcolor{adalotaccent}{\bfseries +\method{} (Ours)}
& 57.4\oursgain{+2.6}
& 54.8\oursgain{+8.2}
& 75.4\oursgain{+3.4}
& 73.0\oursgain{-0.4}
& 74.7\oursgain{+0.1}
& 89.0\oursgain{+0.6}
& 75.8\oursgain{+33.7} \\

\addlinespace
[0.5pt]

& \multirow{3}{*}{Qwen2.5-VL-3B}
& Base
& 59.1 & 59.7 & 62.5 & 79.1 & 76.1 & 88.2 & 17.2 \\
& & \hspace{0.6em}\bfseries +LoT
& 61.7\lotgain{+2.6}
& 61.8\lotgain{+2.1}
& 66.4\lotgain{+3.9}
& 79.5\lotgain{+0.4}
& 79.7\lotgain{+3.6}
& 89.0\lotgain{+0.8}
& 40.5\lotgain{+23.3} \\
& & \hspace{0.6em}\textcolor{adalotaccent}{\bfseries +\method{} (Ours)}
& 62.4\oursgain{+3.3}
& 62.6\oursgain{+2.9}
& 67.0\oursgain{+4.5}
& 79.3\oursgain{+0.2}
& 80.2\oursgain{+4.1}
& 89.4\oursgain{+1.2}
& 41.7\oursgain{+24.5} \\

\addlinespace
[0.5pt]

& \multirow{3}{*}{Qwen3-VL-4B}
& Base
& 66.7 & 56.0 & 74.7 & 80.7 & 76.0 & 90.1 & 81.9 \\
& & \hspace{0.6em}\bfseries +LoT
& 71.6\lotgain{+4.9}
& 67.0\lotgain{+11.0}
& 76.5\lotgain{+1.8}
& 82.3\lotgain{+1.6}
& 77.4\lotgain{+1.4}
& 90.2\lotgain{+0.1}
& 81.2\lotgain{-0.7} \\
& & \hspace{0.6em}\textcolor{adalotaccent}{\bfseries +\method{} (Ours)}
& 72.2\oursgain{+5.5}
& 67.8\oursgain{+11.8}
& 77.1\oursgain{+2.4}
& 82.0\oursgain{+1.3}
& 78.1\oursgain{+2.1}
& 90.5\oursgain{+0.4}
& 82.0\oursgain{+0.1} \\

\addlinespace
[0.5pt]

& \multirow{3}{*}{InternVL3.5-4B}
& Base
& 66.7 & 73.8 & 75.5 & 85.9 & 82.9 & 89.0 & 80.1 \\
& & \hspace{0.6em}\bfseries +LoT
& 67.3\lotgain{+0.6}
& 76.4\lotgain{+2.6}
& 76.0\lotgain{+0.5}
& 85.4\lotgain{-0.5}
& 82.5\lotgain{-0.4}
& 86.4\lotgain{-2.6}
& 73.0\lotgain{-7.1} \\
& & \hspace{0.6em}\textcolor{adalotaccent}{\bfseries +\method{} (Ours)}
& 68.0\oursgain{+1.3}
& 77.1\oursgain{+3.3}
& 76.6\oursgain{+1.1}
& 85.2\oursgain{-0.7}
& 83.1\oursgain{+0.2}
& 87.0\oursgain{-2.0}
& 74.2\oursgain{-5.9} \\

\midrule

\multirow{12}{*}{\rotatebox
[origin=c]
{90}{Medium}}
& \multirow{3}{*}{Qwen2-VL-7B}
& Base
& 62.7 & 52.4 & 78.0 & 81.6 & 80.0 & 70.3 & 33.7 \\
& & \hspace{0.6em}\bfseries +LoT
& 64.8\lotgain{+2.1}
& 56.0\lotgain{+3.6}
& 79.3\lotgain{+1.3}
& 82.0\lotgain{+0.4}
& 81.3\lotgain{+1.3}
& 89.1\lotgain{+18.8}
& 47.3\lotgain{+13.6} \\
& & \hspace{0.6em}\textcolor{adalotaccent}{\bfseries +\method{} (Ours)}
& 65.5\oursgain{+2.8}
& 56.8\oursgain{+4.4}
& 79.8\oursgain{+1.8}
& 81.8\oursgain{+0.2}
& 81.9\oursgain{+1.9}
& 89.4\oursgain{+19.1}
& 48.6\oursgain{+14.9} \\

\addlinespace
[0.5pt]

& \multirow{3}{*}{Qwen2.5-VL-7B}
& Base
& 65.0 & 57.1 & 75.7 & 77.0 & 84.9 & 87.4 & 62.0 \\
& & \hspace{0.6em}\bfseries +LoT
& 67.5\lotgain{+2.5}
& 61.3\lotgain{+4.2}
& 77.9\lotgain{+2.2}
& 79.4\lotgain{+2.4}
& 84.3\lotgain{-0.6}
& 87.4\lotgain{+0.0}
& 62.6\lotgain{+0.6} \\
& & \hspace{0.6em}\textcolor{adalotaccent}{\bfseries +\method{} (Ours)}
& 68.1\oursgain{+3.1}
& 62.0\oursgain{+4.9}
& 78.4\oursgain{+2.7}
& 79.1\oursgain{+2.1}
& 85.0\oursgain{+0.1}
& 87.8\oursgain{+0.4}
& 63.4\oursgain{+1.4} \\

\addlinespace
[0.5pt]

& \multirow{3}{*}{Qwen3-VL-8B}
& Base
& 66.8 & 60.7 & 76.9 & 82.0 & 79.5 & 89.2 & 75.9 \\
& & \hspace{0.6em}\bfseries +LoT
& 67.5\lotgain{+0.7}
& 61.3\lotgain{+0.6}
& 77.9\lotgain{+1.0}
& 79.4\lotgain{-2.6}
& 84.3\lotgain{+4.8}
& 89.2\lotgain{+0.0}
& 62.6\lotgain{-13.3} \\
& & \hspace{0.6em}\textcolor{adalotaccent}{\bfseries +\method{} (Ours)}
& 68.2\oursgain{+1.4}
& 62.0\oursgain{+1.3}
& 78.5\oursgain{+1.6}
& 79.1\oursgain{-2.9}
& 84.9\oursgain{+5.4}
& 89.5\oursgain{+0.3}
& 63.8\oursgain{-12.1} \\

\addlinespace
[0.5pt]

& \multirow{3}{*}{InternVL3.5-8B}
& Base
& 64.7 & 73.3 & 77.2 & 84.3 & 84.3 & 86.3 & 80.4 \\
& & \hspace{0.6em}\bfseries +LoT
& 65.9\lotgain{+1.2}
& 69.6\lotgain{-3.7}
& 78.0\lotgain{+0.8}
& 79.7\lotgain{-4.6}
& 83.0\lotgain{-1.3}
& 87.3\lotgain{+1.0}
& 80.8\lotgain{+0.4} \\
& & \hspace{0.6em}\textcolor{adalotaccent}{\bfseries +\method{} (Ours)}
& 66.5\oursgain{+1.8}
& 70.4\oursgain{-2.9}
& 78.6\oursgain{+1.4}
& 79.5\oursgain{-4.8}
& 83.6\oursgain{-0.7}
& 87.7\oursgain{+1.4}
& 81.4\oursgain{+1.0} \\

\midrule

\multirow{3}{*}{\rotatebox
[origin=c]
{90}{Large}}
& \multirow{3}{*}{Qwen2.5-VL-32B}
& Base
& 65.6 & 55.5 & 72.4 & 35.7 & 78.7 & 88.8 & 89.4 \\
& & \hspace{0.6em}\bfseries +LoT
& 67.1\lotgain{+1.5}
& 65.4\lotgain{+9.9}
& 74.0\lotgain{+1.6}
& 48.2\lotgain{+12.5}
& 79.0\lotgain{+0.3}
& 89.0\lotgain{+0.2}
& 89.0\lotgain{-0.4} \\
& & \hspace{0.6em}\textcolor{adalotaccent}{\bfseries +\method{} (Ours)}
& 67.8\oursgain{+2.2}
& 66.1\oursgain{+10.6}
& 74.6\oursgain{+2.2}
& 50.0\oursgain{+14.3}
& 78.8\oursgain{+0.1}
& 89.3\oursgain{+0.5}
& 89.6\oursgain{+0.2} \\

\bottomrule
\end{tabular}

\caption{Performance on standard MLLM benchmarks using visual evidence only. Base denotes the same checkpoint without evidence highlighting. Colored values in parentheses indicate absolute changes relative to Base.}
\label{tab:standard_main}
\end{table*}

Figure~3 further illustrates how AREA allocates evidence on representative KB-VQA examples. Across questions involving biological, historical, and entity-specific knowledge, AREA localizes the query-relevant visual content while highlighting the retrieved sentences that directly support the answer. The selected evidence is compact and varies across examples, rather than following a uniform sentence budget or crop scale. This behavior is important for KB-VQA, where retrieved passages commonly contain facts that are topically related to the question but insufficient for deriving the answer.

The examples also show that visual and textual evidence need not contribute equally to every prediction. In some cases, the image primarily identifies the queried entity, while the retrieved context supplies the answer-specific fact; in others, accurate visual localization is necessary to disambiguate among several plausible entities before consulting the context. By allocating the two modalities independently, AREA can emphasize the modality that is currently underused without unnecessarily modifying the other. These qualitative results complement the quantitative improvements in Table~1 and suggest that AREA benefits KB-VQA by making short cross-modal evidence chains easier for the frozen model to follow.

\subsection{Generalization to Standard Multimodal Benchmarks}

Table~2 evaluates AREA on standard multimodal benchmarks without retrieved textual context. In this setting, text scaling, text gating, and causal evidence refresh are disabled, while sink filtering, entropy-calibrated crop scaling, and visual intervention gating remain active. Across the 63 checkpoint--benchmark pairs, AREA improves over LoT by approximately 0.55 points on average. The improvement is observed across different backbone families and parameter scales, indicating that adaptive visual allocation is not limited to retrieval-augmented settings or a particular model architecture.

The most consistent gains appear on V-Star and RealWorldQA. V-Star frequently requires recognizing small objects, subtle attributes, or fine-grained spatial cues, for which a question-conditioned crop can increase the relative prominence of the answer-bearing region. RealWorldQA contains more diverse scenes, but its questions often still identify a particular object or local relationship. In such cases, the visual gate is useful because it can retain the original image when broader context is already necessary, rather than applying a compact crop unconditionally. This may explain why AREA improves consistently over LoT on both locally focused and more heterogeneous real-world questions.

AREA also produces stable, although more moderate, gains on TextVQA and OCRBench. Visual text may occupy only a small part of an image, making localization beneficial, but the required words can also be distributed across multiple signs, lines, or document regions. Entropy-calibrated scaling provides a compromise: concentrated relevance leads to a tighter crop, whereas dispersed relevance preserves a larger visual extent. Nevertheless, evidence allocation alone cannot resolve recognition errors caused by small fonts, low resolution, or difficult layouts, which likely limits the magnitude of the improvement on these benchmarks.

The clearest exception is ChartQA, where AREA remains approximately comparable to LoT and occasionally performs slightly worse. Chart questions often require jointly interpreting axes, legends, categories, and spatially separated numerical values. Cropping around the most salient region may therefore remove structural information needed to relate these components, even when the localized content itself is relevant. The improvement obtained with Qwen2.5-VL-32B suggests that stronger models may recover missing global relationships more effectively, but the overall pattern indicates that localized evidence highlighting is less suitable for tasks whose reasoning depends on the complete visual layout.

On the hallucination-oriented benchmarks, AREA consistently improves over LoT, with the largest additional gains observed on AMBER-D. Sink filtering can prevent highly activated but semantically uninformative visual tokens from dominating localization, while intervention gating avoids altering inputs for which the original visual evidence is already sufficiently represented. However, highlighting still degrades some strong backbones relative to their Base performance on AMBER-D, suggesting that adaptive allocation mitigates rather than fully eliminates the risks of visual intervention. POPE shows smaller gains, likely because several checkpoints already achieve high accuracy and leave limited room for improvement. Overall, these results support the central premise of AREA: visual highlighting is most effective when its necessity and spatial extent are determined by the evidence distribution of each input, rather than fixed in advance.

\FloatBarrier

\section{Conclusion}

We investigated evidence utilization in frozen multimodal large language models and showed that fixed highlighting policies cannot adapt to diverse evidence demands. 
We proposed Adaptive Relevance-guided Evidence Allocation (AREA), a training-free framework that dynamically determines whether to intervene, how much evidence to expose, and when to refresh textual evidence during generation. 
By leveraging relevance signals from a single probe pass, AREA performs adaptive evidence scaling and modality-specific intervention without modifying model parameters. 
Experiments across KB-VQA and general multimodal benchmarks demonstrate that AREA consistently improves over fixed highlighting strategies, especially when answers rely on compact cross-modal evidence. 
These results highlight the importance of adaptive evidence presentation for improving MLLM inference.
\FloatBarrier

\bibliography{area_arxiv}

\clearpage
\pdfpagewidth=210mm
\pdfpageheight=297mm
\paperwidth=210mm
\paperheight=297mm
\setlength{\textwidth}{190mm}
\setlength{\textheight}{277mm}
\setlength{\oddsidemargin}{\dimexpr10mm-1in\relax}
\setlength{\evensidemargin}{\dimexpr10mm-1in\relax}
\setlength{\topmargin}{\dimexpr10mm-1in\relax}
\setlength{\headheight}{0pt}
\setlength{\headsep}{0pt}
\onecolumn
\makeatletter
\global\@colht\textheight
\global\@colroom\textheight
\global\vsize\textheight
\let\section\arxiv@article@section
\let\subsection\arxiv@article@subsection
\let\normalsize\arxiv@article@normalsize
\let\small\arxiv@article@small
\let\footnotesize\arxiv@article@footnotesize
\let\large\arxiv@article@large
\let\Large\arxiv@article@Large
\let\LARGE\arxiv@article@LARGE
\let\@makecaption\arxiv@article@makecaption
\makeatother
\renewcommand{\rmdefault}{cmr}
\renewcommand{\sfdefault}{cmss}
\renewcommand{\ttdefault}{cmtt}
\normalfont\normalsize
\nonfrenchspacing
\fussy
\raggedbottom
\pagestyle{empty}
\setlength{\parindent}{0pt}
\setlength{\parskip}{0pt}
\setcounter{table}{0}
\setcounter{figure}{0}

\makeatletter
\def\@title{\bfseries Appendix for\\[1mm]
Layers, Sinks, and Scaling: Adaptive Evidence Selection for\\
Multimodal Large Language Models}
\def\@author{}
\def\@date{}
\arxiv@article@maketitle
\makeatother

\thispagestyle{empty}
\vspace{-3em}
\begin{center}
\begin{minipage}{0.94\linewidth}
\small
This standalone appendix accompanies \textit{Layers, Sinks, and Scaling:
Adaptive Evidence Selection for Multimodal Large Language Models}. It contains
three complementary parts. Part~I reports the runtime comparison between LoT
and AREA under a common model and GPU configuration. Part~II isolates the
contributions of AREA's evidence scaling, intervention gating, and textual
refresh components. Part~III presents qualitative comparisons across ten
benchmarks, exposing the predictions and the visual and textual evidence
selected by each method. Together, these results supplement the main paper
with efficiency details, component-level analysis, and instance-level evidence.
\end{minipage}
\end{center}
\vspace{2mm}
\section*{Part I: Runtime Efficiency}
The table below reports average end-to-end latency per sample. Per-sample time
is used so that the comparison is not confounded by the different sharding
layouts of the recorded full-dataset runs.
\vspace{1mm}
\begin{table}[H]
\centering
\caption{Per-sample inference runtime of LoT and AREA using Qwen2.5-VL-3B-Instruct in BF16 with batch size 1 on an NVIDIA A30 24\,GB GPU per process. Each time is the recorded sum of per-sample elapsed times divided by the number of evaluated samples. The timed region includes retrieval, evidence processing, and answer generation, but excludes model/dataset initialization and metric evaluation. Runtime reduction is $(t_{\mathrm{LoT}}-t_{\mathrm{AREA}})/t_{\mathrm{LoT}}\times100\%$. Positive values mean AREA is faster.}
\label{tab:runtime}
\vspace{2mm}
\begingroup
\small
\renewcommand{\arraystretch}{1.25}
\setlength{\tabcolsep}{8pt}
\begin{tabular}{lrrrr}
\toprule
Dataset & Samples & LoT (s/sample) & AREA (s/sample) & Faster (\%) \\
\midrule
E-VQA & 5,750 & 1.363 & \textbf{1.018} & \textbf{25.3\%} \\
ViQuAE & 1,257 & 1.927 & \textbf{1.272} & \textbf{34.0\%} \\
RealWorldQA & 765 & 1.019 & \textbf{0.874} & \textbf{14.2\%} \\
VstarBench & 191 & 1.225 & \textbf{0.894} & \textbf{27.1\%} \\
TextVQA & 5,000 & 1.250 & \textbf{0.973} & \textbf{22.2\%} \\
ChartQA & 2,500 & 1.177 & \textbf{0.909} & \textbf{22.8\%} \\
OCRBench & 1,000 & 1.170 & \textbf{0.878} & \textbf{25.0\%} \\
POPE & 9,000 & 0.992 & \textbf{0.922} & \textbf{7.0\%} \\
AMBER-D & 14,216 & 1.164 & \textbf{1.049} & \textbf{9.8\%} \\
\bottomrule
\end{tabular}
\endgroup
\vspace{2mm}
\begin{minipage}{0.94\linewidth}
\end{minipage}
\end{table}
\clearpage
\section*{Part II: Algorithm Ablation}
AREA makes three complementary evidence-allocation decisions: how much evidence
to expose through Entropy-Calibrated Evidence Scaling, whether to intervene
through Modality-Specific Intervention Gating, and when to revisit textual
evidence through Causal Text-Evidence Refresh. Table~\ref{tab:ablation}
isolates these decisions using one-factor-at-a-time ablations. All variants use
the same Qwen2.5-VL-3B-Instruct checkpoint, retrieved entities (when applicable),
dataset splits, prompts, decoding settings, and evaluators; only the component
named by each column is removed. Causal Text-Evidence Refresh is not applicable
to vision-only benchmarks because they provide no retrieved textual context.
\vspace{1mm}
\begin{table}[H]
\centering
\caption{Component ablation of AREA. The full AREA column reproduces the
Qwen2.5-VL-3B-Instruct results reported in the main tables.
\textbf{w/o Scaling} fixes the textual evidence budget to one sentence and the
visual crop scale to $\beta=2$; \textbf{w/o Gating} applies the candidate
intervention to every eligible modality; and \textbf{w/o Refresh} disables
generation-time textual evidence refresh. All other settings are held fixed.
Higher is better for every metric, and the best value in each row is bold.}
\label{tab:ablation}
\vspace{2mm}
\begingroup
\small
\renewcommand{\arraystretch}{1.22}
\setlength{\tabcolsep}{7pt}
\begin{tabular}{llcccc}
\toprule
Dataset & Metric & \textbf{AREA} & \textbf{w/o Scaling} &
\textbf{w/o Gating} & \textbf{w/o Refresh} \\
\midrule
\multicolumn{6}{l}{\textit{Retrieval-augmented benchmarks}} \\
E-VQA & All Accuracy & \textbf{31.0} & 30.5 & 30.7 & 30.8 \\
InfoSeek & All Accuracy & \textbf{25.6} & 25.3 & 25.2 & 25.4 \\
ViQuAE & Exact Match & \textbf{28.5} & 27.9 & 28.0 & 28.2 \\
\cmidrule(lr){1-6}
Retrieval Avg. & Mean & \textbf{28.4} & 27.9 & 28.0 & 28.1 \\
\addlinespace[2pt]
\multicolumn{6}{l}{\textit{Vision-only benchmarks}} \\
RealWorldQA & Accuracy & \textbf{62.4} & 61.9 & 61.8 & -- \\
V-Star & Accuracy & \textbf{62.6} & 61.7 & 62.0 & -- \\
TextVQA & VQA Accuracy & \textbf{67.0} & 66.3 & 66.5 & -- \\
ChartQA & Relaxed Accuracy & 79.3 & \textbf{79.5} & 78.9 & -- \\
OCRBench & Overall Score & \textbf{80.2} & 79.6 & 79.8 & -- \\
POPE & F1 & \textbf{89.4} & 89.1 & 89.0 & -- \\
AMBER-D & Accuracy & \textbf{41.7} & 41.2 & 40.6 & -- \\
\cmidrule(lr){1-6}
Vision-only Avg. & Mean & \textbf{68.9} & 68.5 & 68.4 & -- \\
\bottomrule
\end{tabular}
\endgroup
\end{table}
The full AREA column serves as the matched reference. Differences from it
quantify the contribution of each decision without changing the underlying
model or retrieval inputs. For vision-only tasks, the scaling and gating
ablations operate on visual evidence alone, while the refresh entry is marked
as not applicable.
\clearpage
\begin{center}
{\Large\bfseries Part III: Qualitative Visualizations}\par
\vspace{1mm}
\begin{minipage}{0.94\linewidth}
\small
The following pages present three instance-level comparisons for each of ten
benchmarks. Each panel reports the question, reference answer, both
predictions, localized visual evidence, and the surrounding retrieved context
with selected sentences highlighted. Vision-only benchmarks explicitly state
that no external retrieval is used.
\end{minipage}
\end{center}
\vspace{1.5mm}
\begin{center}
{\LARGE\bfseries E-VQA: AREA vs. LoT}\par
\vspace{1mm}
\end{center}
\vspace{1mm}
\includegraphics[width=\linewidth]{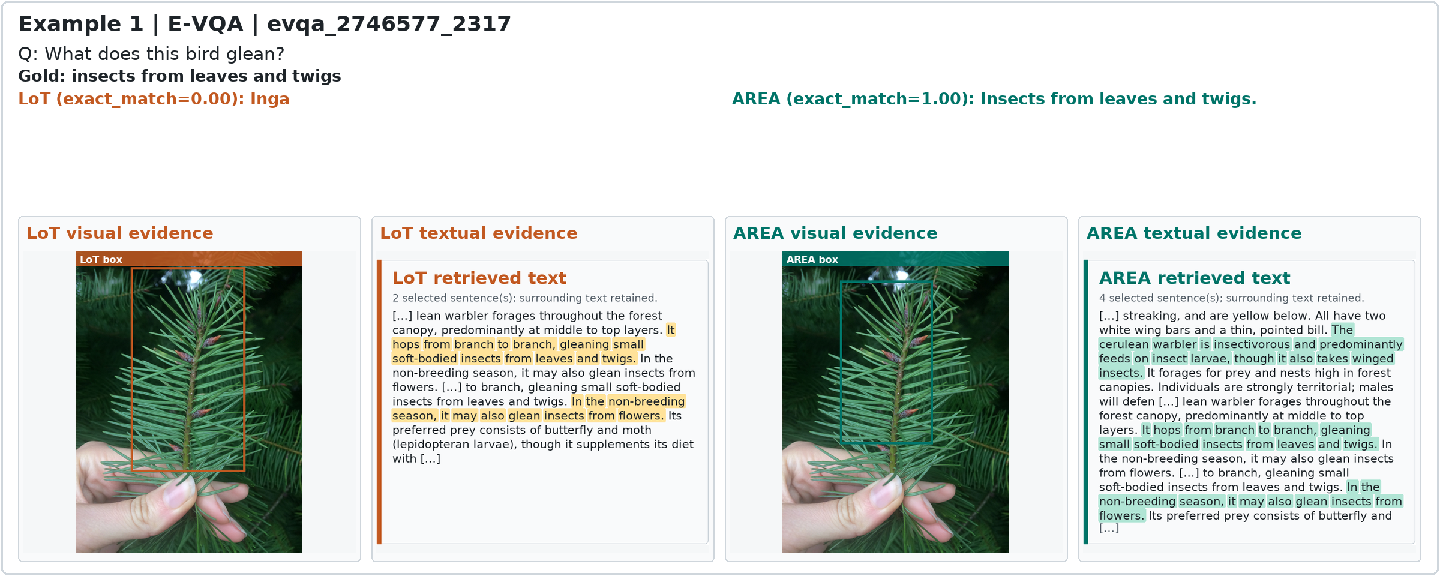}\par
\vspace{1.5mm}
\includegraphics[width=\linewidth]{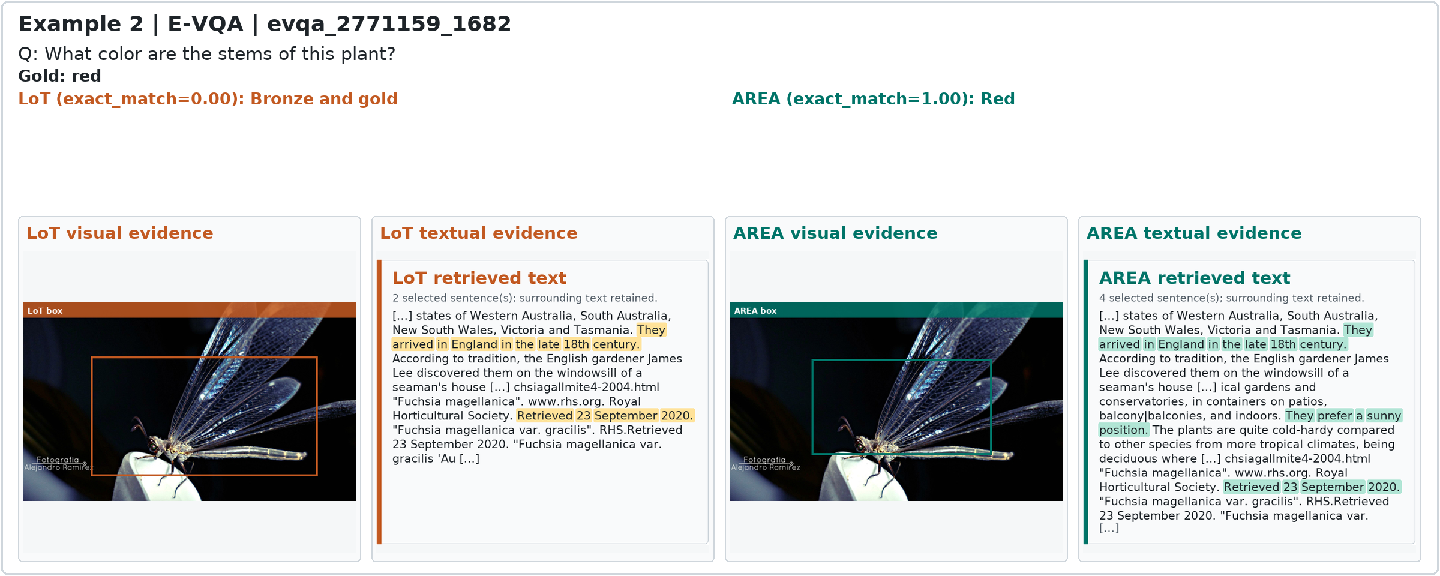}\par
\vspace{1.5mm}
\includegraphics[width=\linewidth]{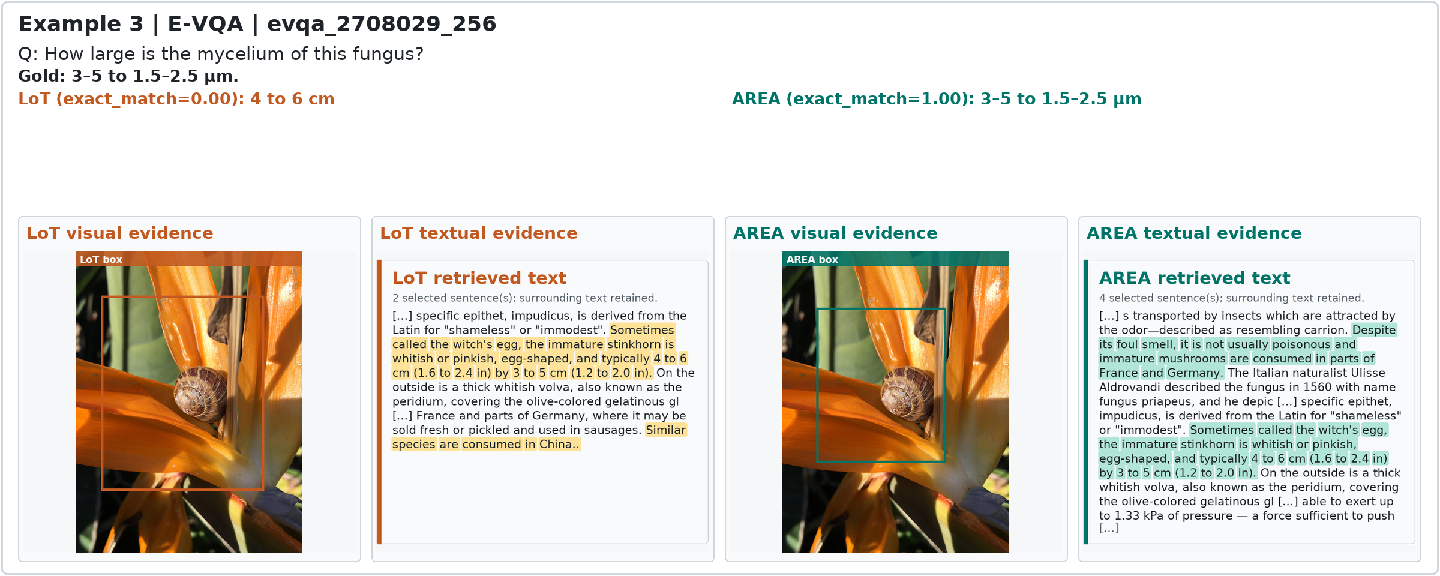}\par
\clearpage
\begin{center}
{\LARGE\bfseries InfoSeek: AREA vs. LoT}\par
\vspace{1mm}
\end{center}
\vspace{1mm}
\includegraphics[width=\linewidth]{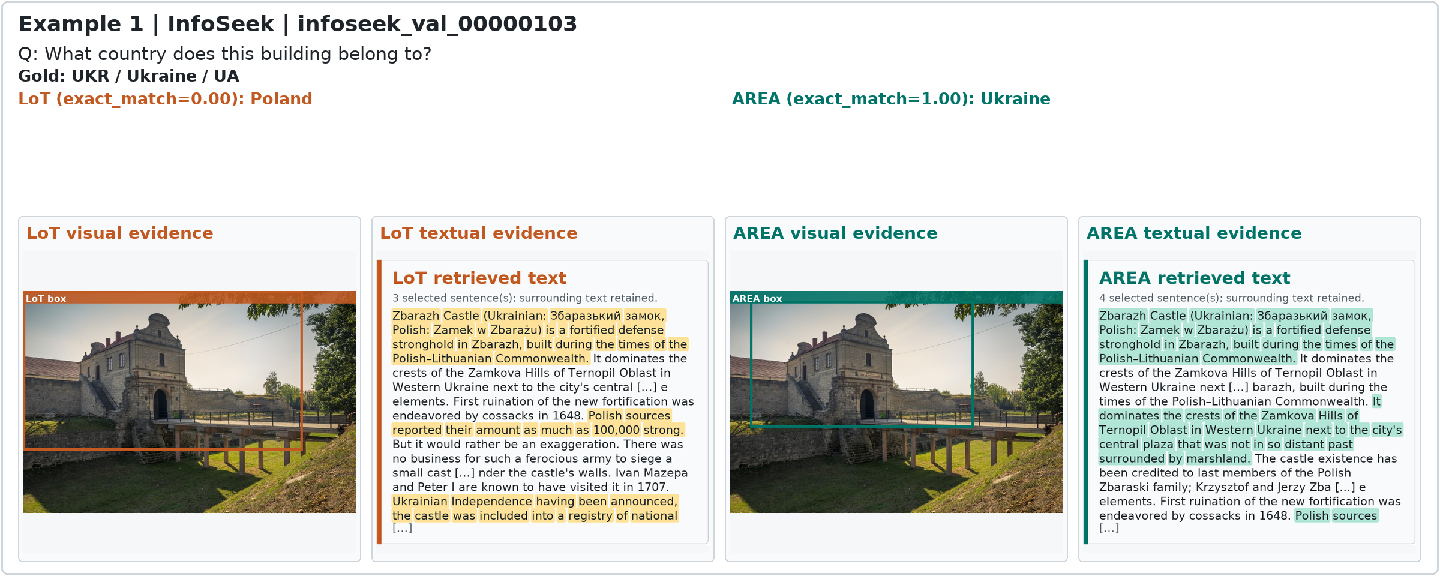}\par
\vspace{1.5mm}
\includegraphics[width=\linewidth]{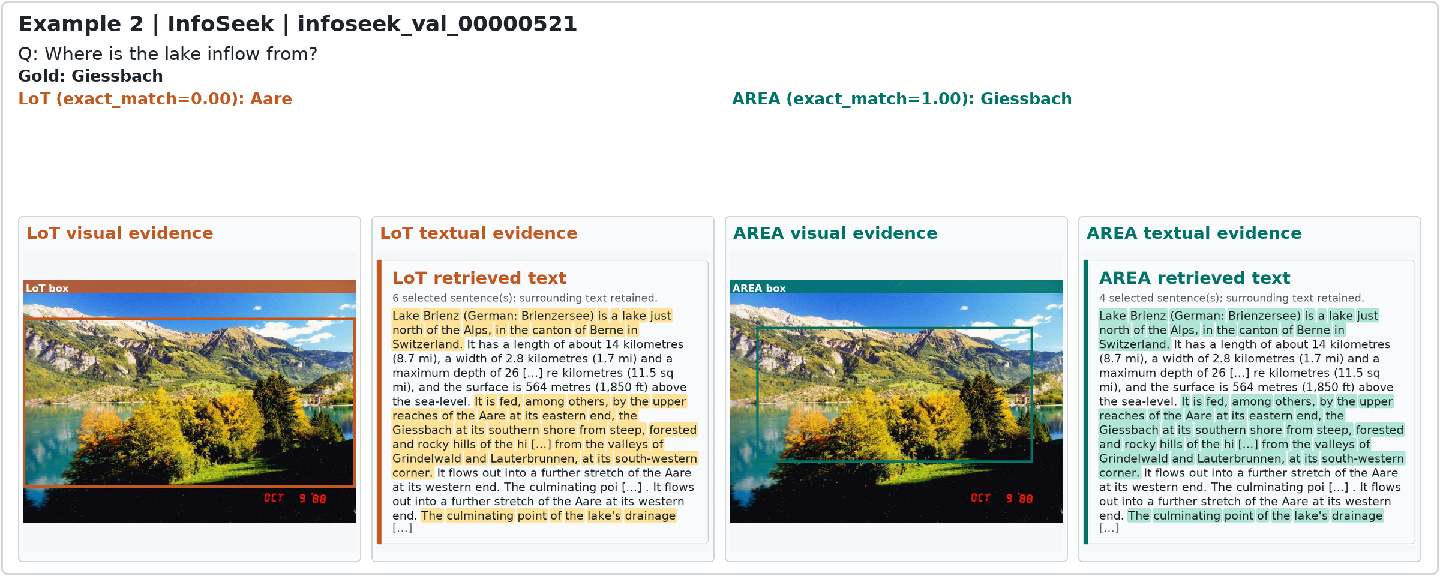}\par
\vspace{1.5mm}
\includegraphics[width=\linewidth]{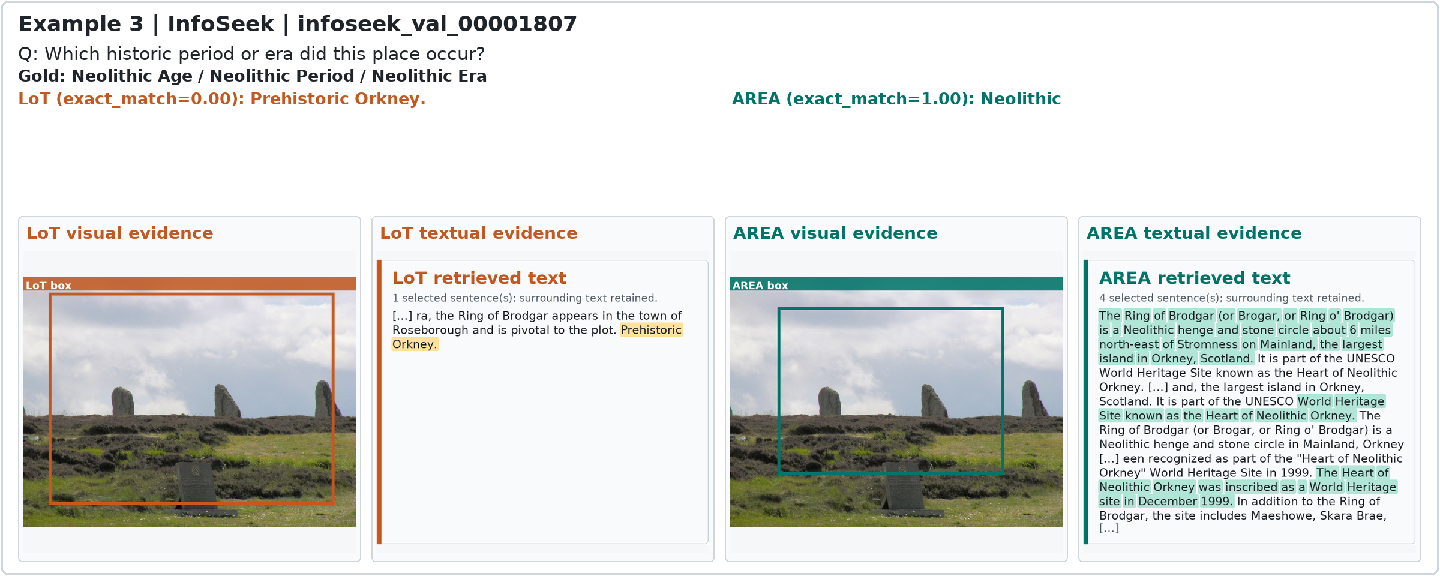}\par
\clearpage
\begin{center}
{\LARGE\bfseries ViQuAE: AREA vs. LoT}\par
\vspace{1mm}
\end{center}
\vspace{1mm}
\includegraphics[width=\linewidth]{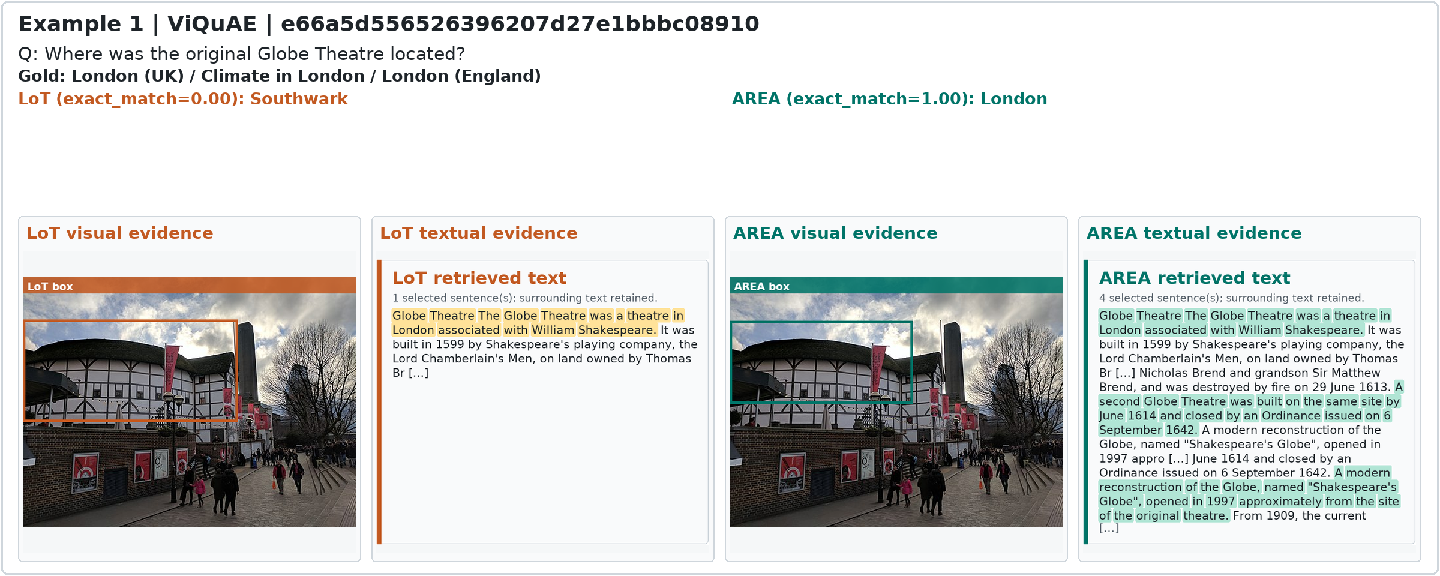}\par
\vspace{1.5mm}
\includegraphics[width=\linewidth]{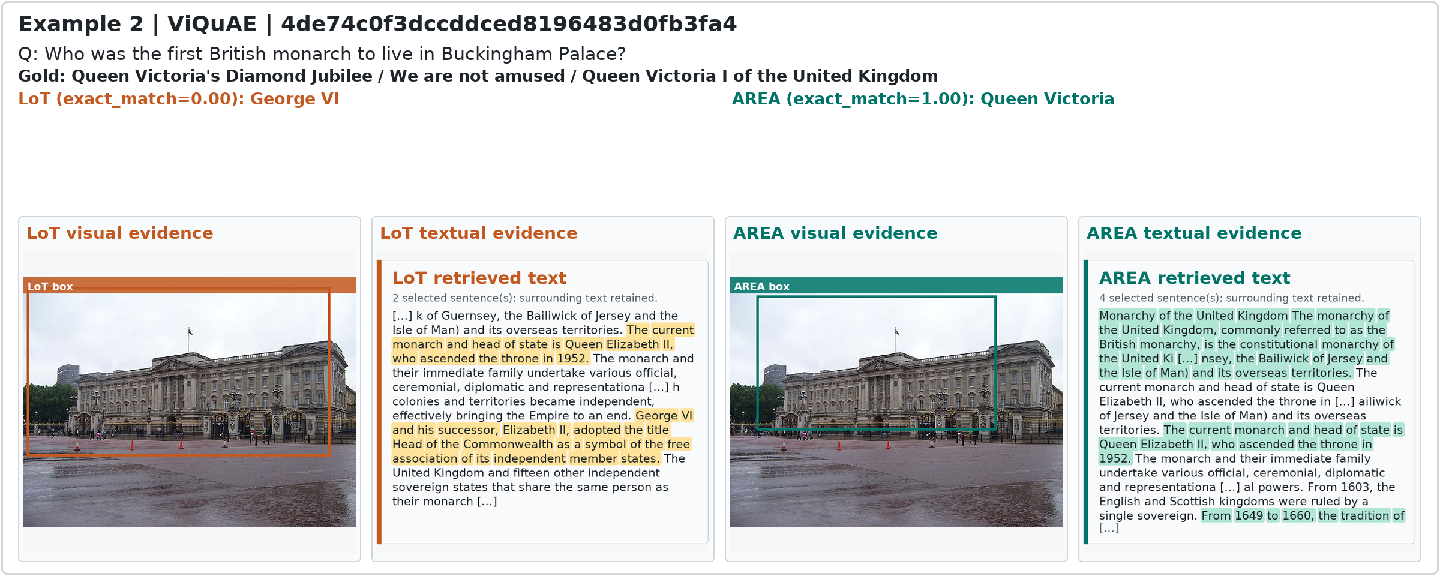}\par
\vspace{1.5mm}
\includegraphics[width=\linewidth]{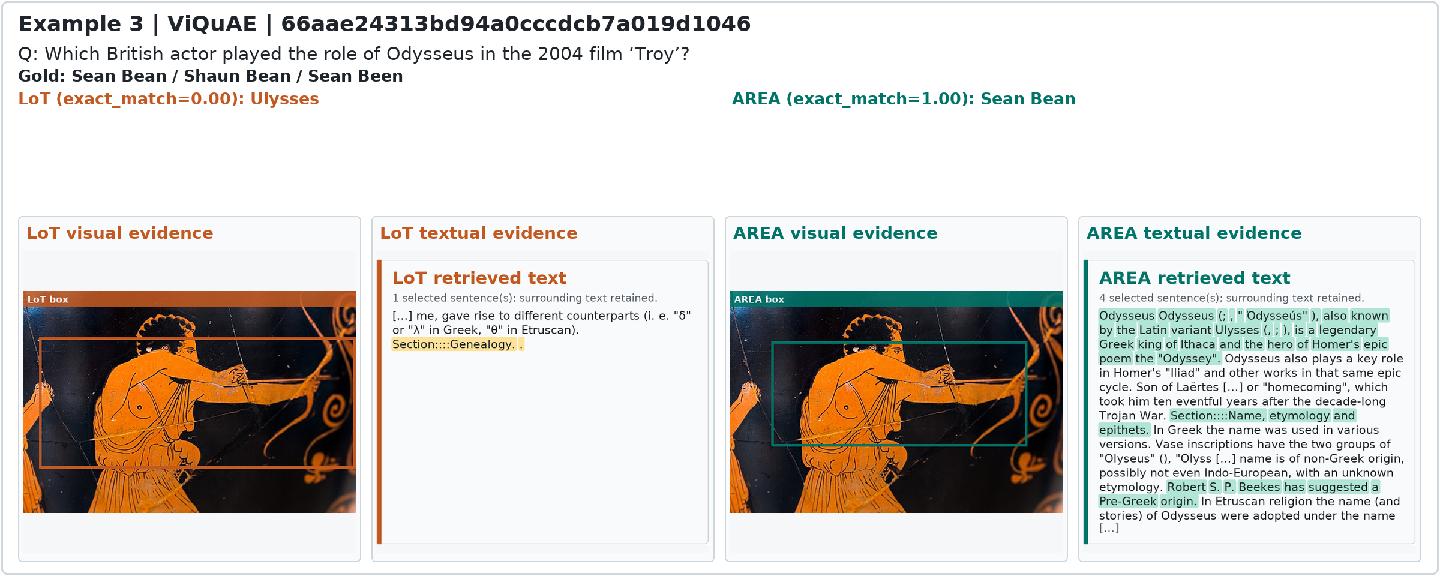}\par
\clearpage
\begin{center}
{\LARGE\bfseries RealWorldQA: AREA vs. LoT}\par
\vspace{1mm}
 
\end{center}
\vspace{1mm}
\includegraphics[width=\linewidth]{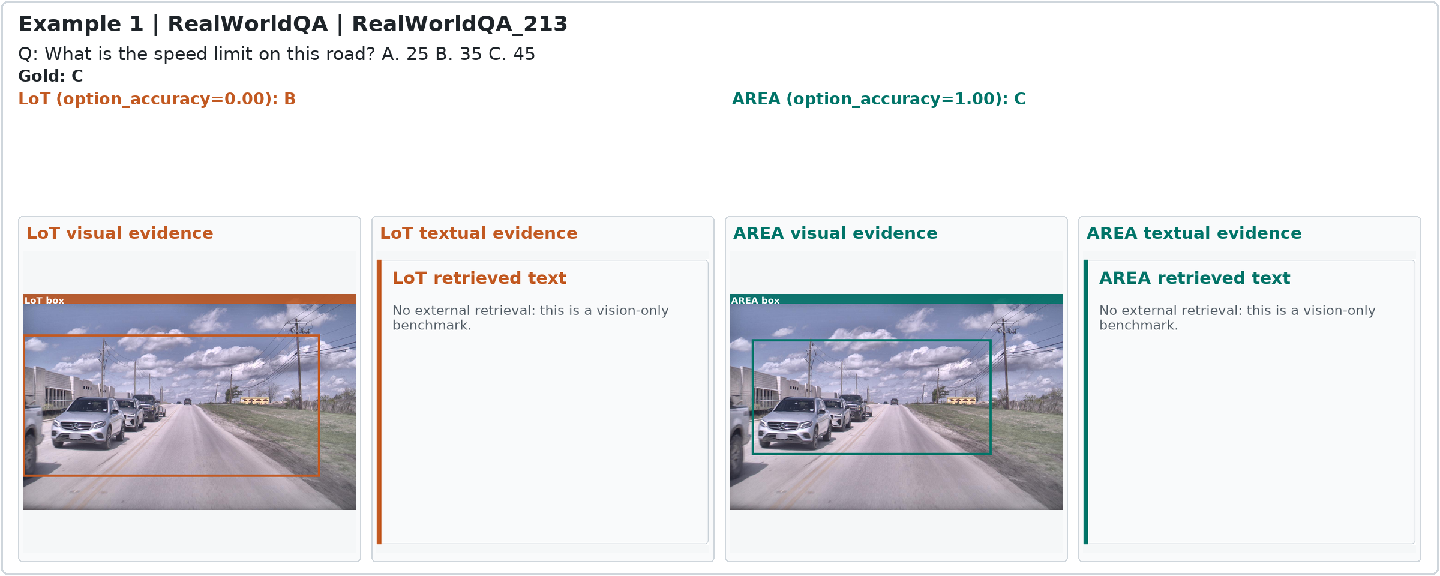}\par
\vspace{1.5mm}
\includegraphics[width=\linewidth]{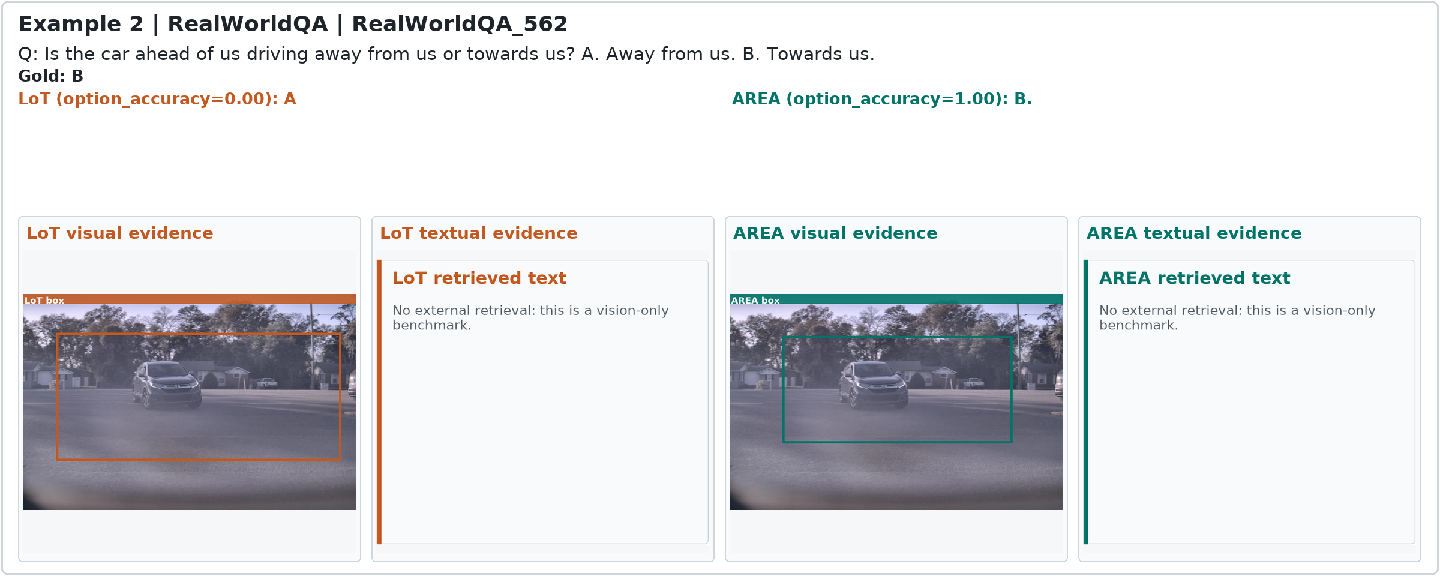}\par
\vspace{1.5mm}
\includegraphics[width=\linewidth]{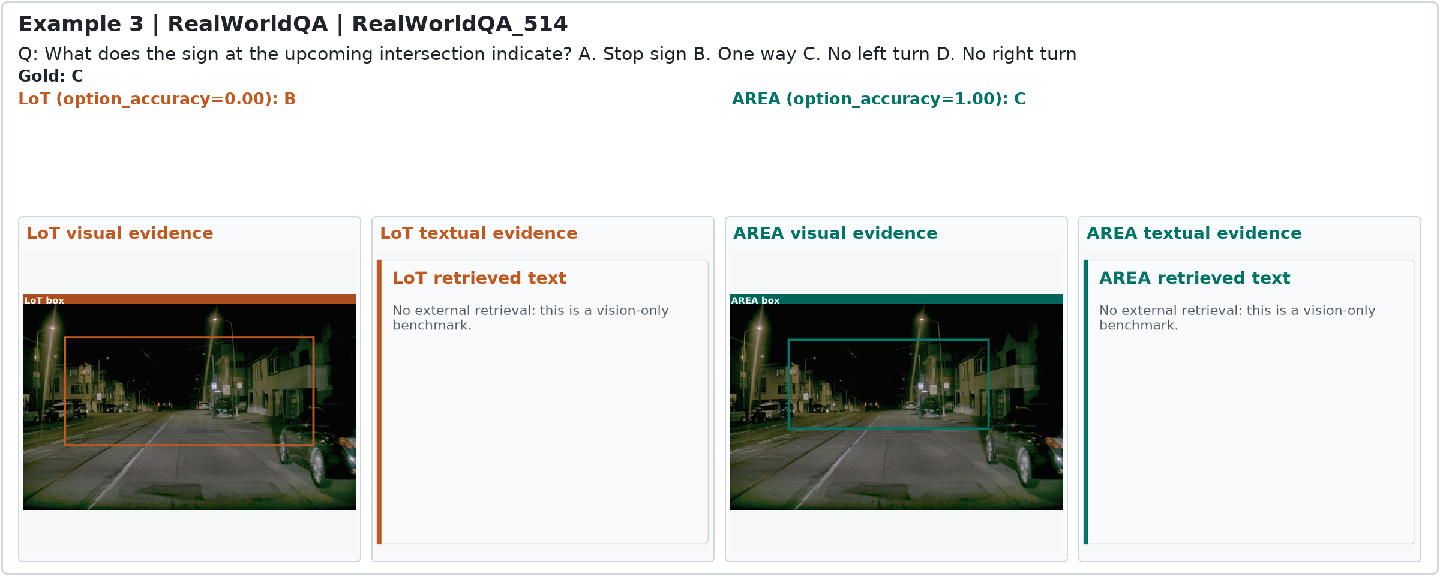}\par
\clearpage
\begin{center}
{\LARGE\bfseries VstarBench: AREA vs. LoT}\par
\vspace{1mm}
 
\end{center}
\vspace{1mm}
\includegraphics[width=\linewidth]{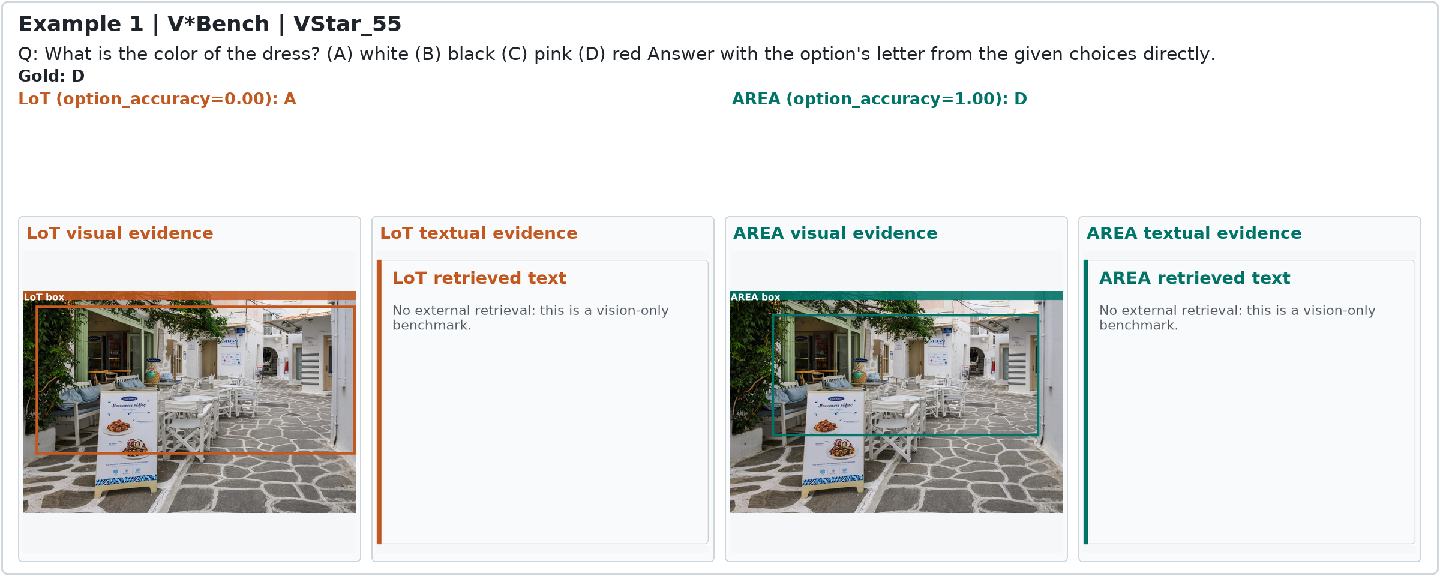}\par
\vspace{1.5mm}
\includegraphics[width=\linewidth]{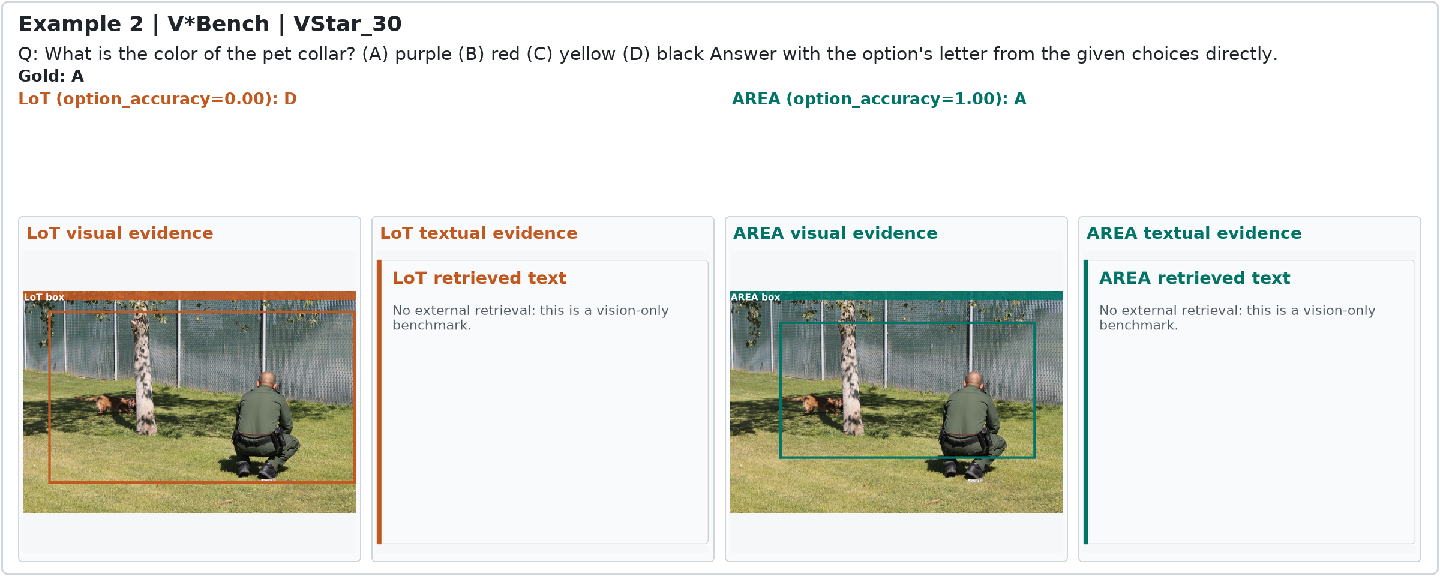}\par
\vspace{1.5mm}
\includegraphics[width=\linewidth]{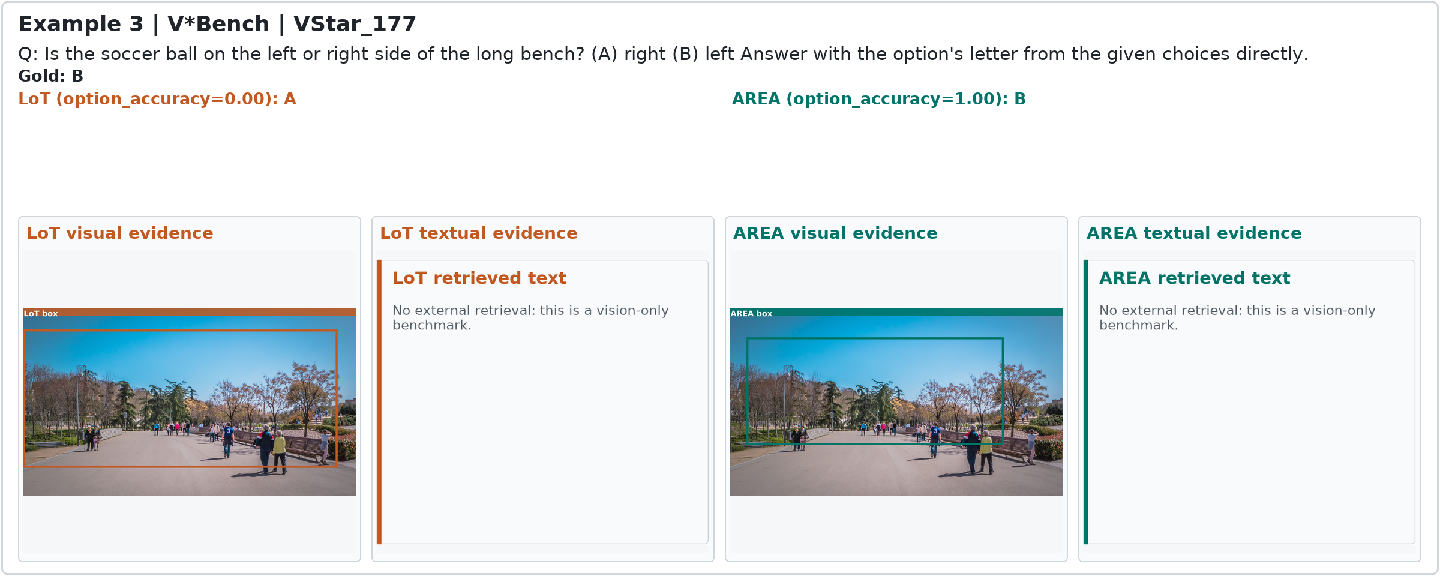}\par
\clearpage
\begin{center}
{\LARGE\bfseries TextVQA: AREA vs. LoT}\par
\vspace{1mm}
 
\end{center}
\vspace{1mm}
\includegraphics[width=\linewidth]{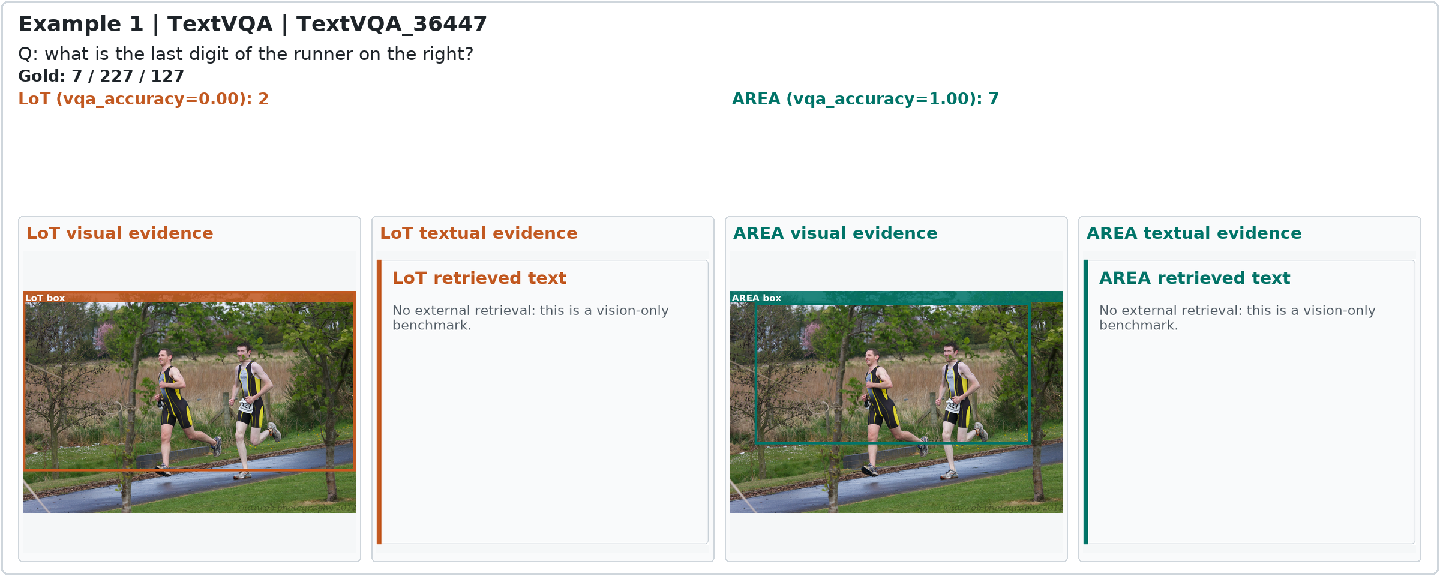}\par
\vspace{1.5mm}
\includegraphics[width=\linewidth]{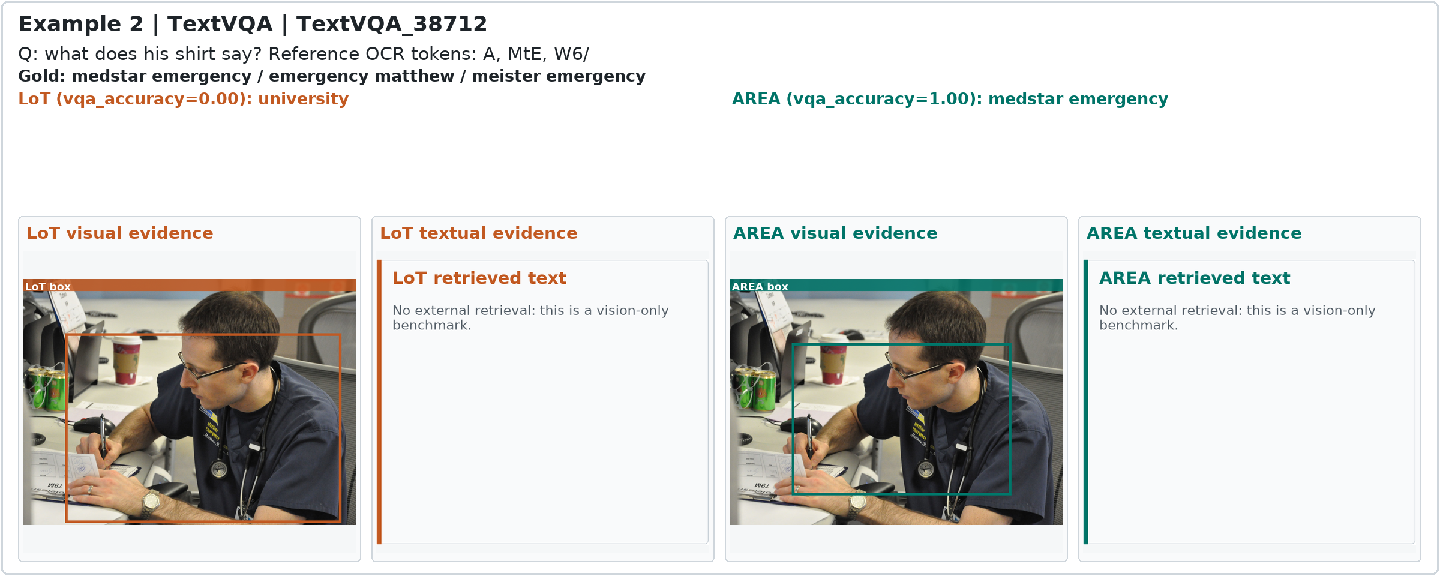}\par
\vspace{1.5mm}
\includegraphics[width=\linewidth]{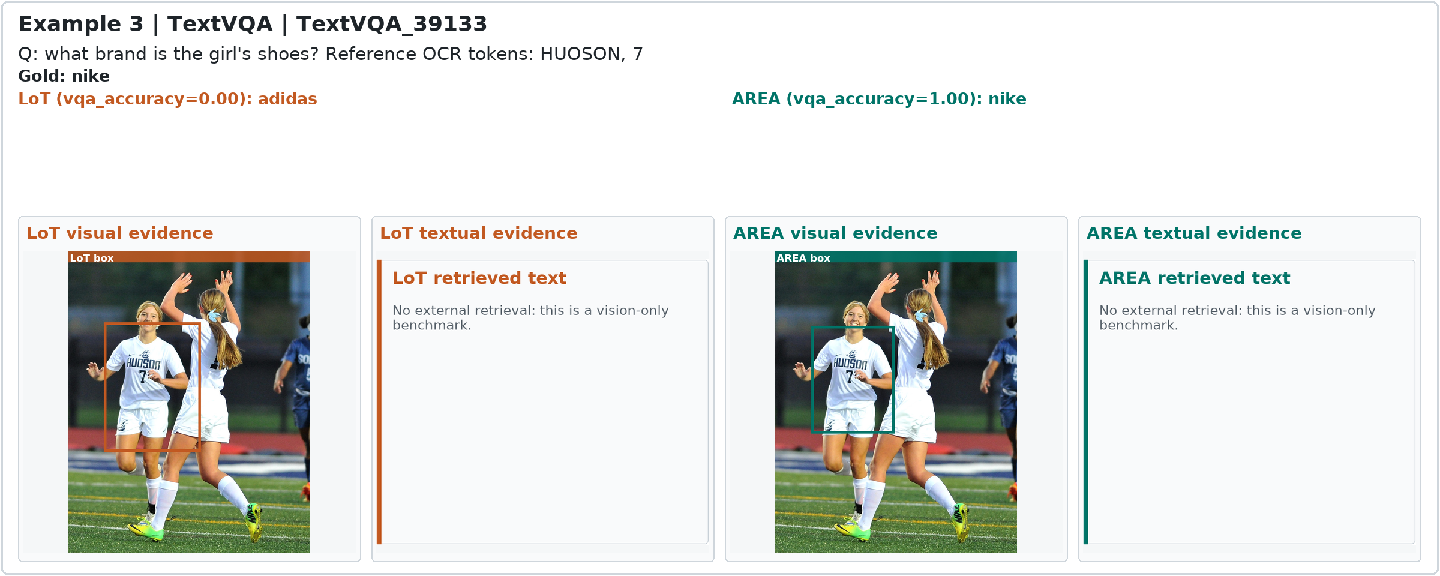}\par
\clearpage
\begin{center}
{\LARGE\bfseries ChartQA: AREA vs. LoT}\par
\vspace{1mm}
 
\end{center}
\vspace{1mm}
\includegraphics[width=\linewidth]{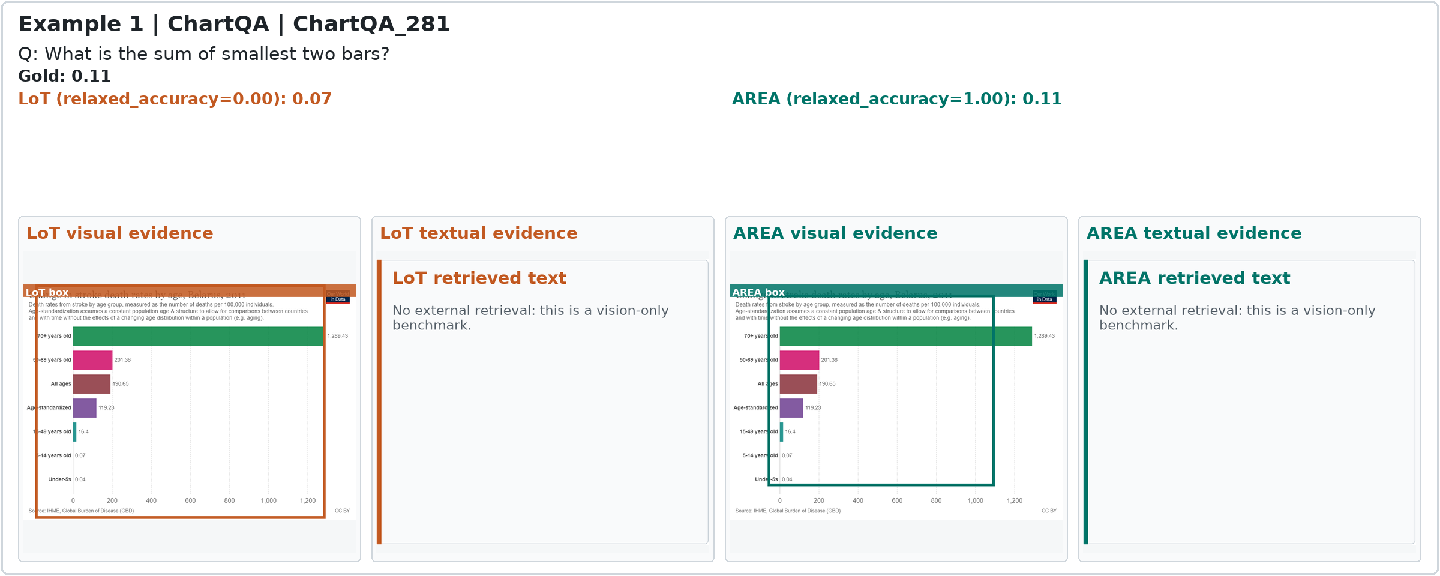}\par
\vspace{1.5mm}
\includegraphics[width=\linewidth]{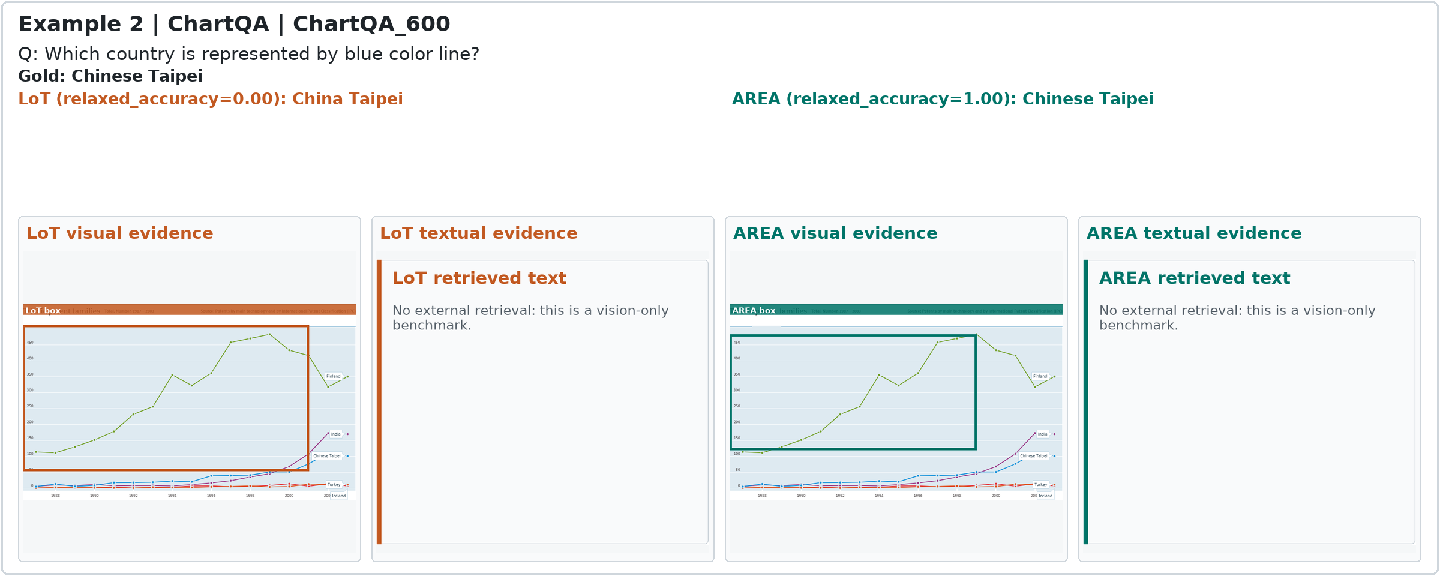}\par
\vspace{1.5mm}
\includegraphics[width=\linewidth]{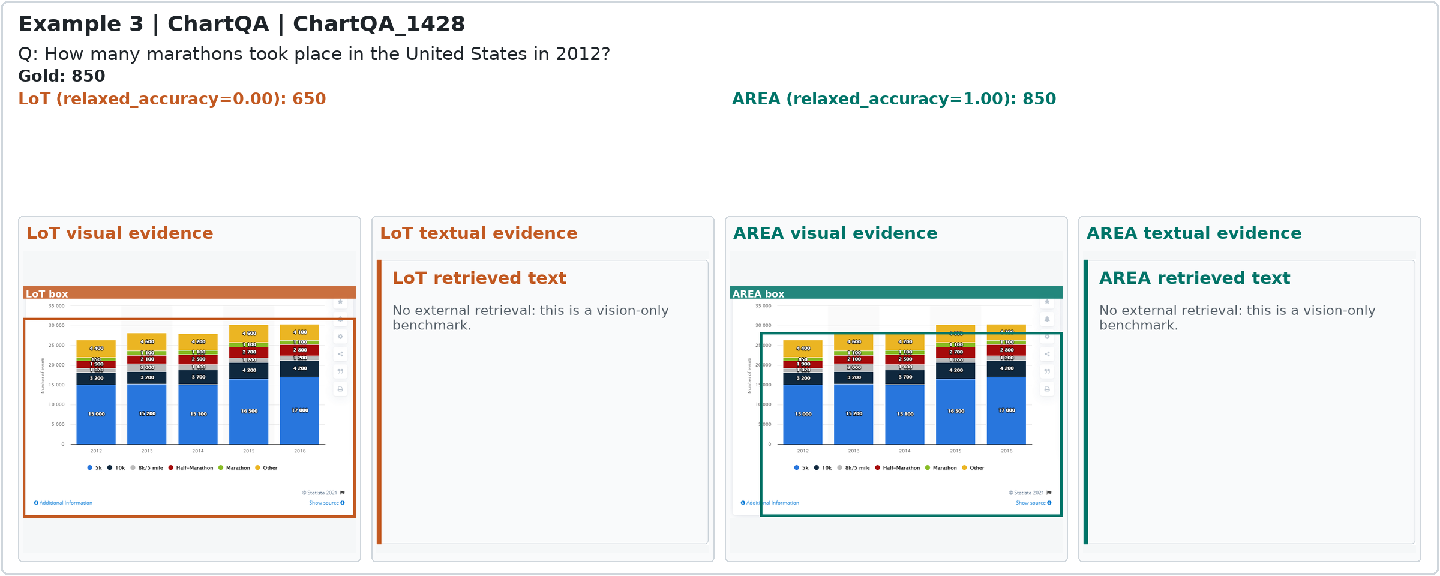}\par
\clearpage
\begin{center}
{\LARGE\bfseries OCRBench: AREA vs. LoT}\par
\vspace{1mm}
 
\end{center}
\vspace{1mm}
\includegraphics[width=\linewidth]{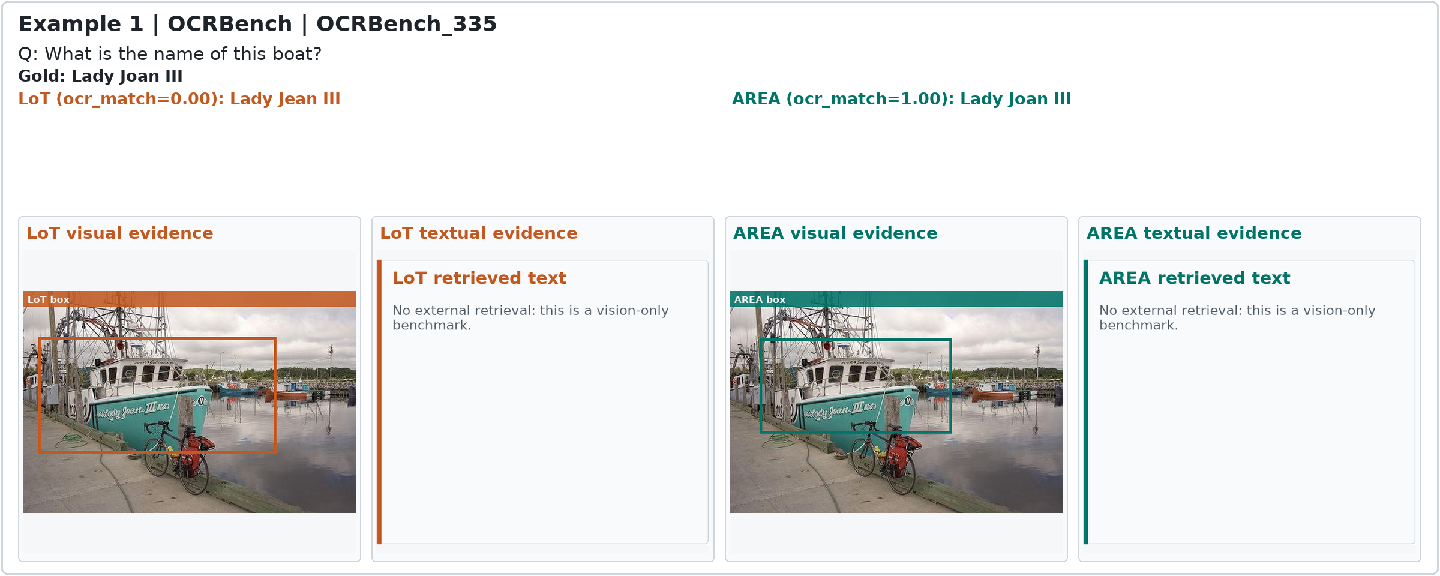}\par
\vspace{1.5mm}
\includegraphics[width=\linewidth]{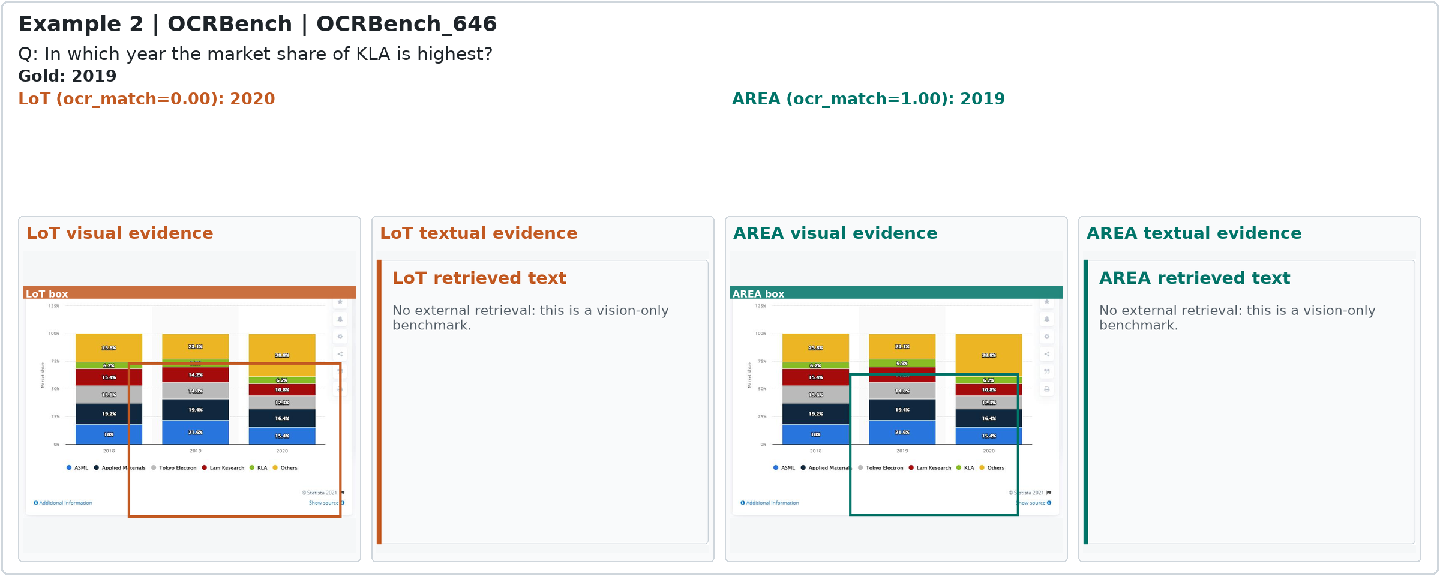}\par
\vspace{1.5mm}
\includegraphics[width=\linewidth]{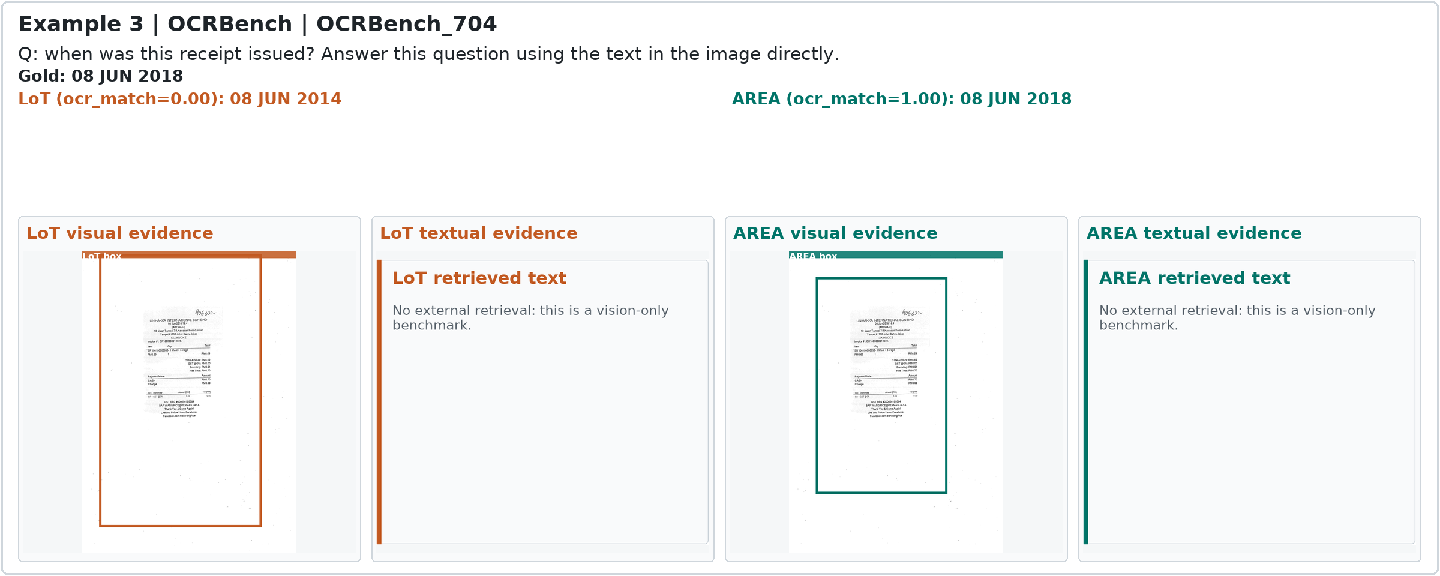}\par
\clearpage
\begin{center}
{\LARGE\bfseries POPE: AREA vs. LoT}\par
\vspace{1mm}
 
\end{center}
\vspace{1mm}
\includegraphics[width=\linewidth]{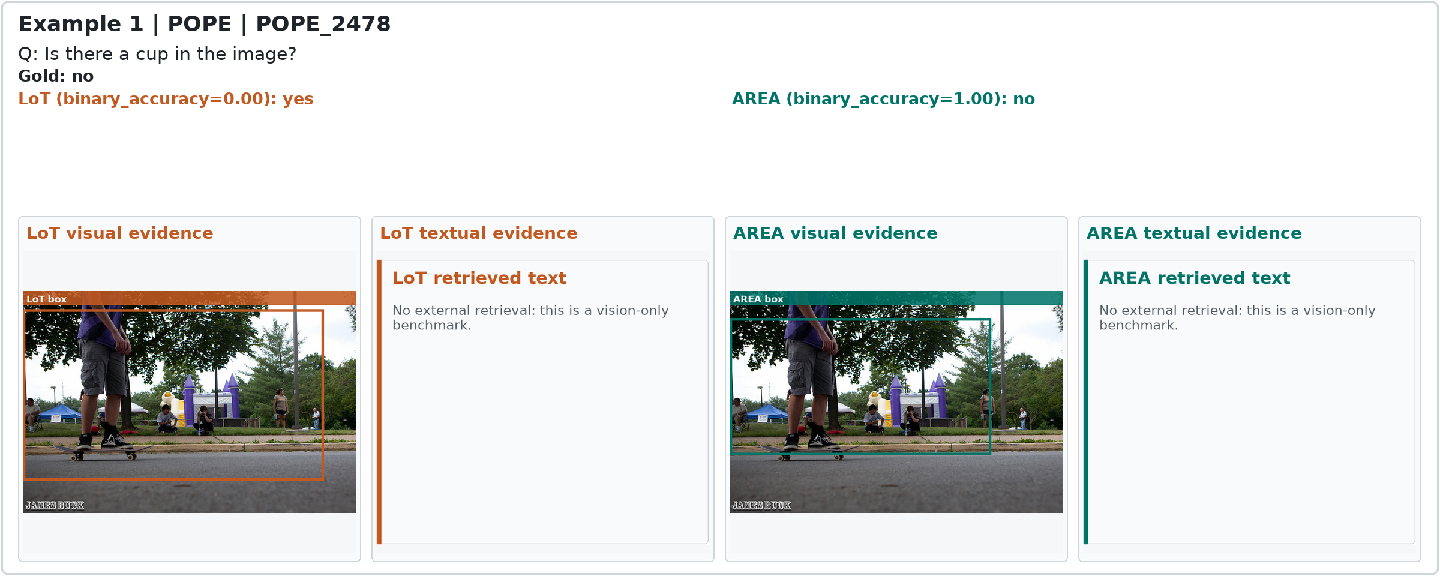}\par
\vspace{1.5mm}
\includegraphics[width=\linewidth]{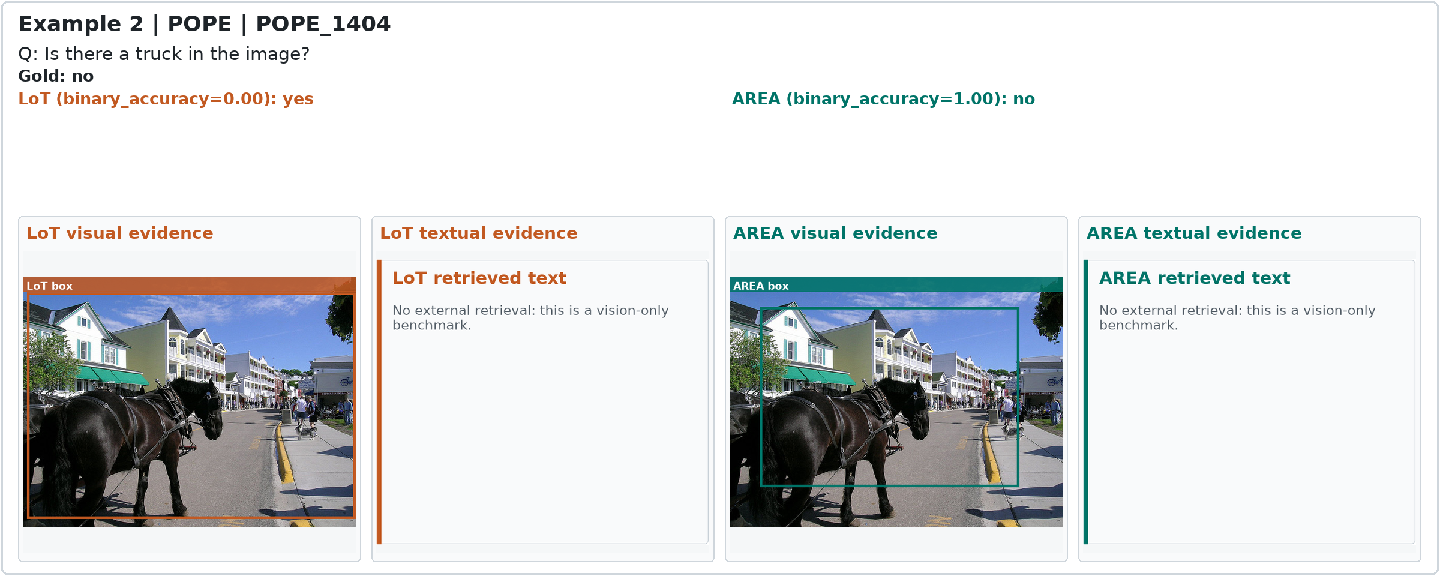}\par
\vspace{1.5mm}
\includegraphics[width=\linewidth]{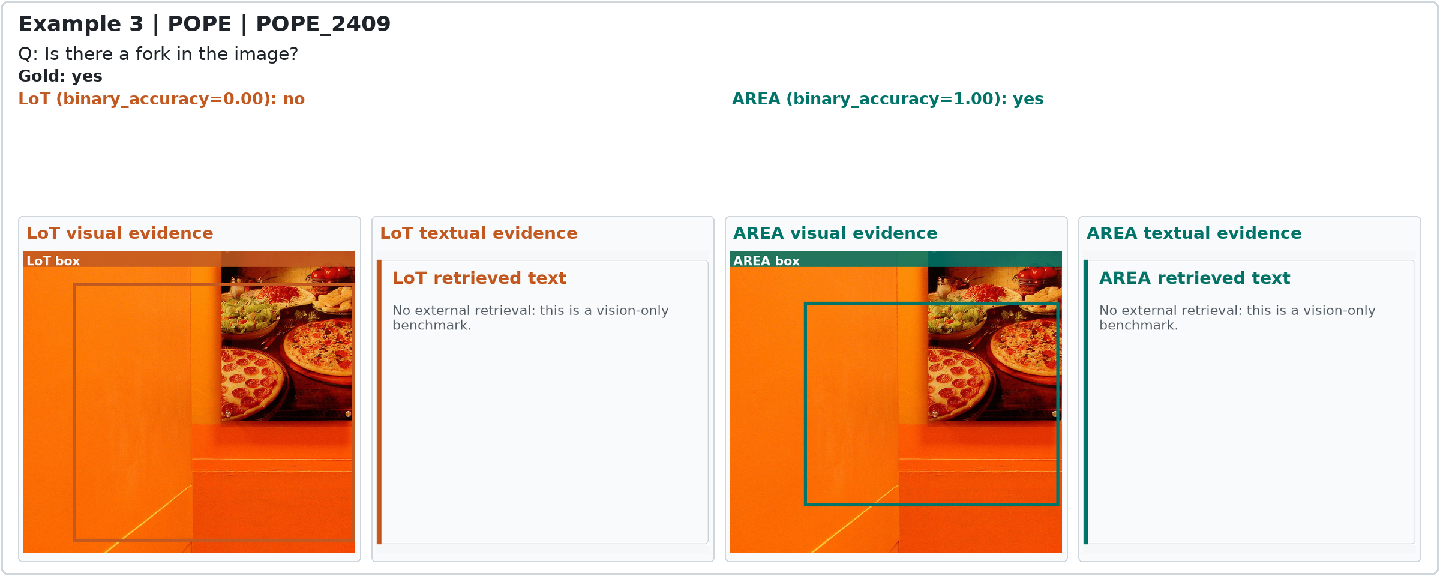}\par
\clearpage
\begin{center}
{\LARGE\bfseries AMBER-D: AREA vs. LoT}\par
\vspace{1mm}
 
\end{center}
\vspace{1mm}
\includegraphics[width=\linewidth]{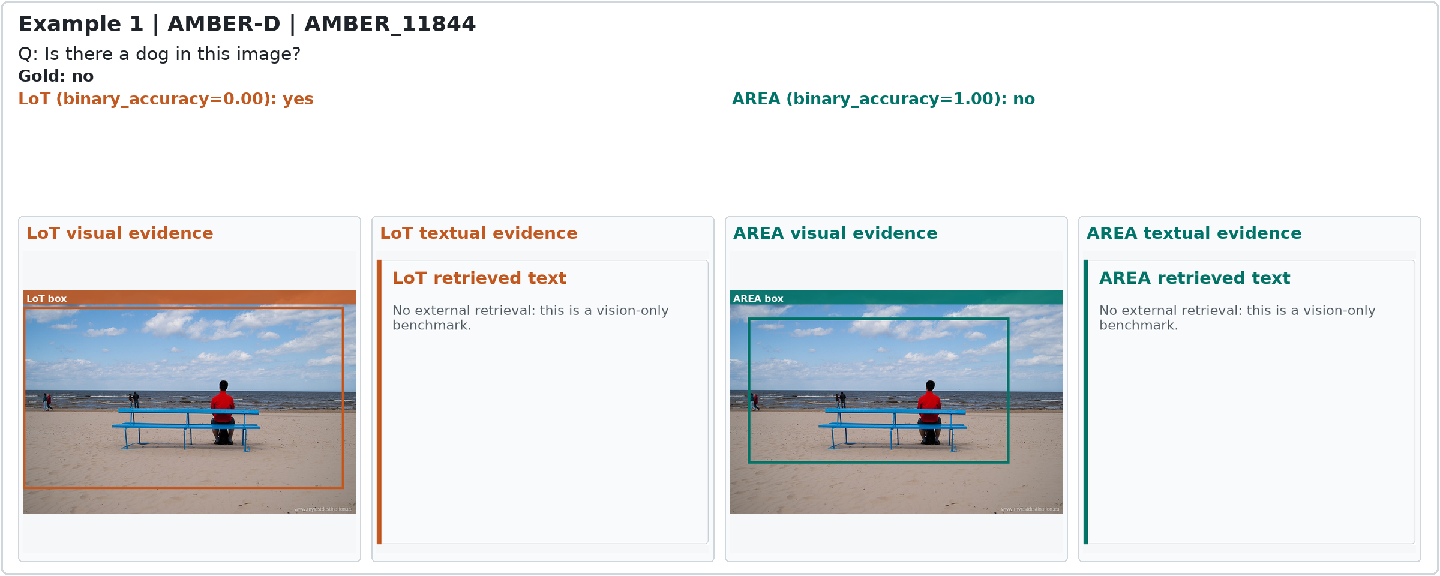}\par
\vspace{1.5mm}
\includegraphics[width=\linewidth]{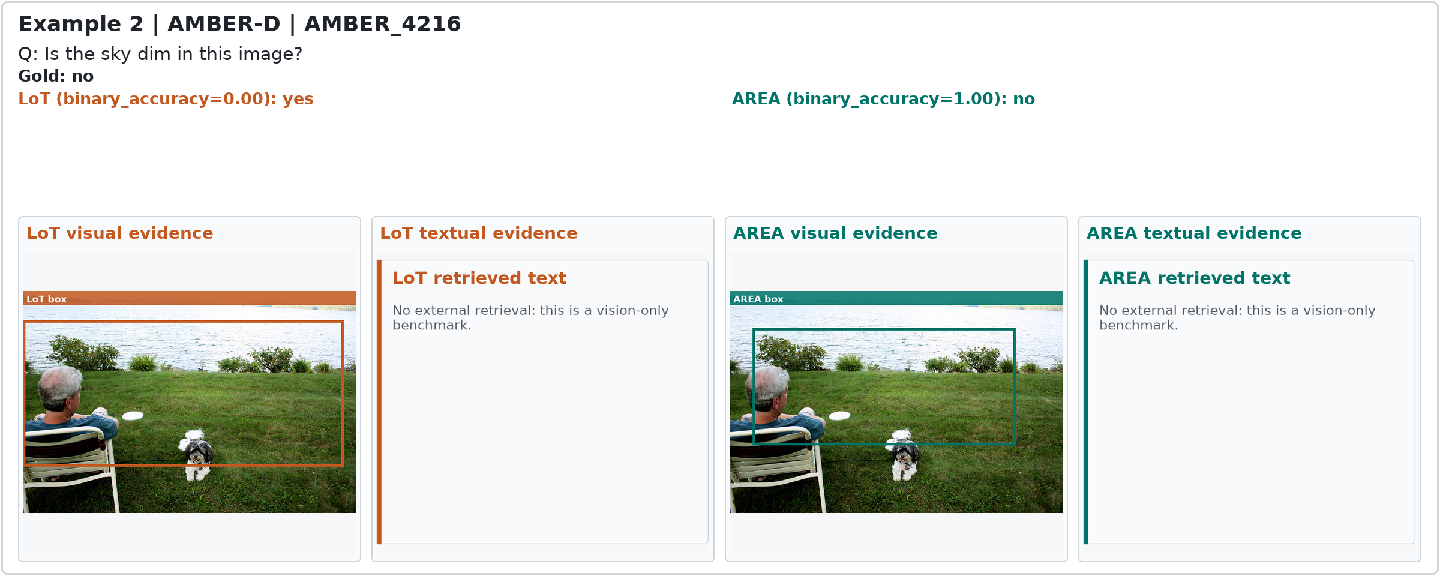}\par
\vspace{1.5mm}
\includegraphics[width=\linewidth]{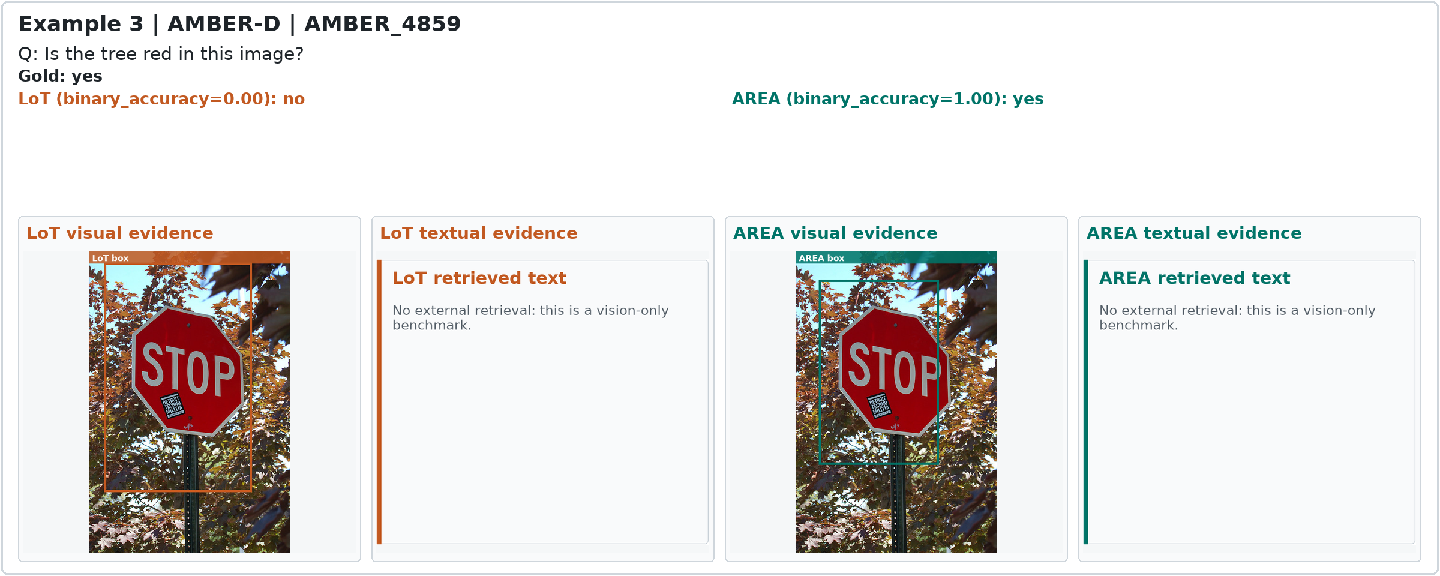}\par

\end{document}